%% file: main.tex
\documentclass[11pt]{article}
\usepackage[T1]{fontenc}
\usepackage{times}
\usepackage[letterpaper,margin=1in]{geometry}
\usepackage[dvipsnames]{xcolor}
\usepackage{amsmath,amssymb,booktabs,graphicx,xspace,url}
\usepackage[numbers,sort&compress]{natbib}
\usepackage[format=plain,labelformat=simple,labelsep=period,font=small]{caption}
\usepackage[font=footnotesize,skip=3pt]{subcaption}
\usepackage[shortlabels,inline]{enumitem}
\input{paper_tables}
\usepackage{algorithm}
\usepackage{algpseudocode}

\newcommand{\appref}[1]{\ref{#1}}
\definecolor{cvprblue}{rgb}{0.21,0.49,0.74}
\definecolor{impf}{HTML}{5B6FD8}
\usepackage[pagebackref,breaklinks,colorlinks,allcolors=cvprblue]{hyperref}
\hypersetup{pdfauthor={Ruiqi Liu, Jinming Li, Mo Wang, Yang Lu, Yanlan Kang},pdftitle={Choosing What to Observe: Budget-Paired Counterfactual Audits for Credible Evaluation of Visual Agents}}
\makeatletter\@secpenalty=0\makeatother
\title{\fontsize{13.6}{17}\selectfont\bfseries Choosing What to Observe:\\Budget-Paired Counterfactual Audits for Credible Evaluation of Visual Agents}
\author{%
Ruiqi Liu$^{1,2}$\qquad Jinming Li$^{1}$\qquad Mo Wang$^{3,4,*}$\qquad Yang Lu$^{1,*}$\qquad Yanlan Kang$^{2,*}$\\[0.7em]
{\normalsize $^{1}$Fudan University\qquad $^{2}$\textbf{\textcolor{impf}{Institute of Medical Philosophy \& Future AI (IMPF-AI)}}}\\
{\normalsize $^{3}$The University of Warwick\qquad $^{4}$Southern University of Science and Technology}\\[0.3em]
{\small $^{*}$Corresponding authors}}
\date{}
\begin{document}
\maketitle
\vspace{-1.6em}
\begin{center}
\raisebox{-.5\height}{\includegraphics[height=0.50in]{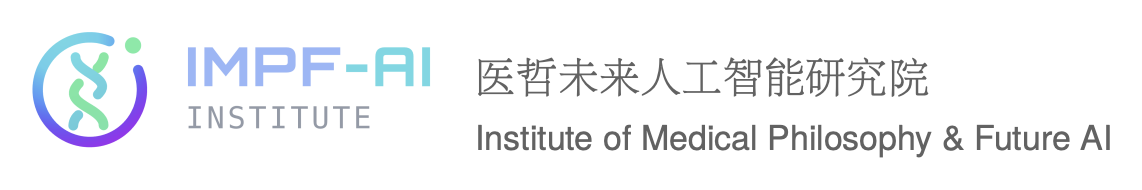}}\hspace{2.4em}%
\raisebox{-.5\height}{\includegraphics[height=0.40in]{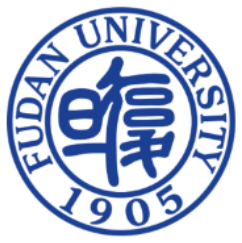}}\hspace{1.8em}%
\raisebox{-.5\height}{\includegraphics[height=0.40in]{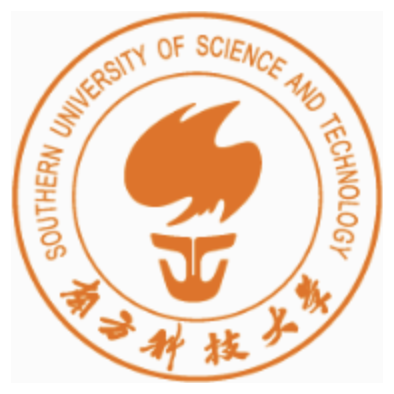}}\hspace{1.8em}%
\raisebox{-.5\height}{\includegraphics[height=0.36in]{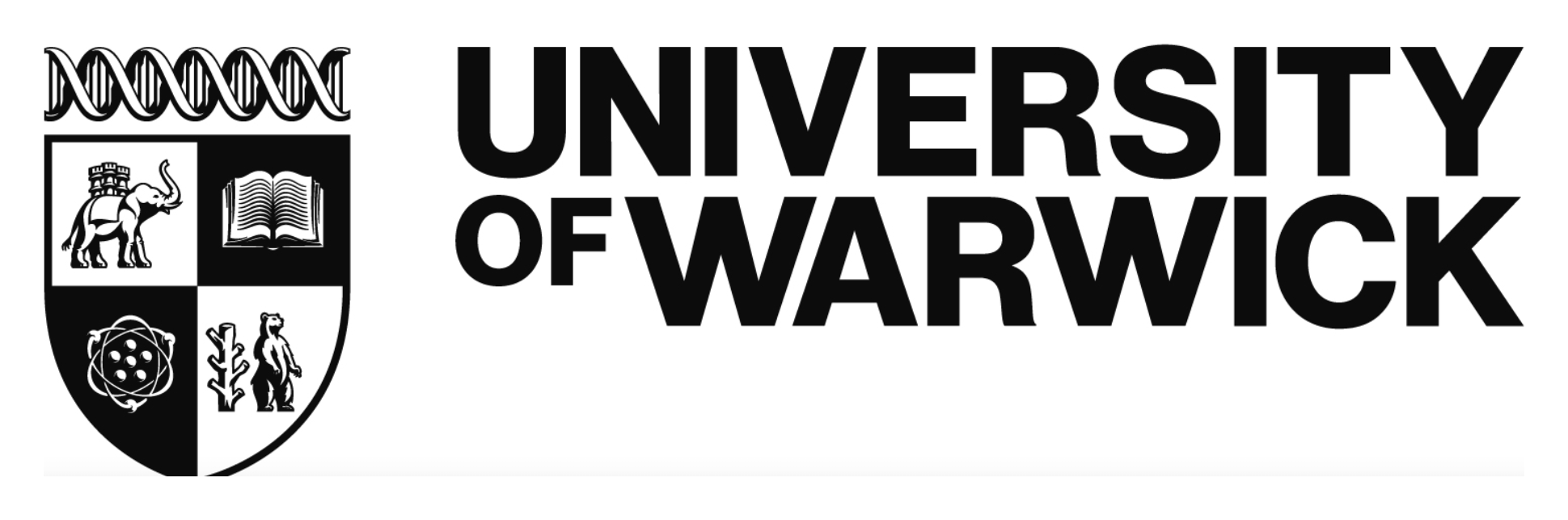}}
\end{center}
\vspace{0.4em}
\begin{center}
\captionsetup{type=figure}
\includegraphics[width=\linewidth]{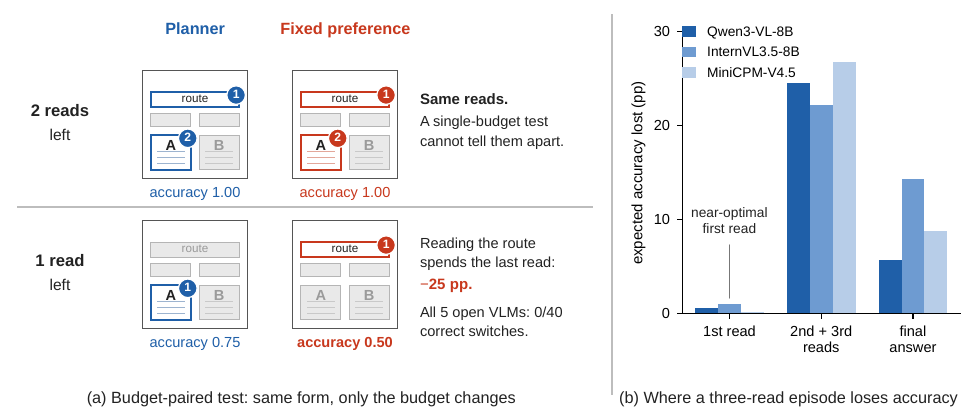}
\captionof{figure}{\textbf{Budget-paired counterfactual audit.} \textbf{(a)}~A form whose route field says which price table (A or B) holds the answer; numbered circles give the order of reads. With two reads left, a horizon-aware planner and a policy that always reads the route first make identical choices, so a single-budget test cannot separate them. With one read left, the route is worth less than either table (expected accuracy 0.50 versus 0.75), and only the planner switches. All five open-weight VLMs keep reading the route: 0/40 optimal switches. \textbf{(b)}~Expected accuracy lost at each step of 160 three-read episodes per model: first reads are near-optimal (0.18--0.96~pp), while the second and third reads lose 22.20--26.76~pp.}
\label{fig:visual-example}
\end{center}

\begin{abstract}
\noindent Visual agents that read dense documents must ration a finite observation budget across candidate regions, yet answer accuracy conflates where an agent looks, how it adapts to the reads that remain, and how it answers once observation stops. We introduce \emph{budget-paired counterfactual audits}, an evaluation protocol that holds public evidence and region bindings fixed while perturbing only the remaining budget or declared task parameters, and credits every acquired history with exact continuation values. The resulting accounting decomposes lost expected accuracy into stepwise acquisition loss and terminal decision loss, separating planning from fixed-role heuristics. Across forty paired routing anchors, five open-weight VLMs make 0/40 optimal branch-to-selector switches and incur 12.50--15.62 percentage points (pp) of anchor loss, a cost that low non-anchor averages (0.59--0.85~pp) conceal. In three-read hierarchies, first reads are nearly optimal (0.18--0.96~pp loss), whereas the second and third reads lose 22.20--26.76~pp. On thirty FUNSD forms, Qwen3-VL-8B's one-crop selection gain over a shared uniform reader falls from 24.44~pp under curated references to an inconclusive 10.56~pp under the original references, and its two-crop pipeline trails native whole-image reading (70.0\% versus 83.3\%) while using twice the input tokens. Credible evaluation of visual agents therefore requires auditing complete observation histories against matched whole-image baselines, with transparent reference scoring.
\end{abstract}

\section{Introduction}
Visual agents often work on dense documents from a low-resolution overview that exposes the layout but blurs fine text. Selective high-resolution crops recover local detail without processing every pixel~\cite{vstar,hrbench,fovea}, which turns observation into a sequential decision problem whenever reads are limited.

Consider an invoice whose routing field indicates which of two price tables applies. With two crops remaining, reading the routing field first directs the second crop to the active table. With one crop left, the same action exhausts the budget without revealing a price, and opening a table directly is strictly better. An observation is worth what it enables before the reads run out, not what it reveals in isolation.

Answer accuracy on isolated test cases cannot see this distinction. Our key idea is to audit \emph{budget-paired counterfactuals}: decisions whose public evidence and region bindings stay identical while only the remaining budget or the action-value boundary changes (Fig.~\ref{fig:visual-example}a). Such pairs separate policies that look identical under a single budget. A fixed-selector heuristic and a horizon-aware planner both open the routing field at $b=2$; only the planner opens a branch at $b=1$.

Detecting a behavioral gap is half of the problem; an evaluator must also price it across the whole rollout. Under an explicit task model, we compute exact finite-horizon continuation values and apply a performance-difference identity that attributes lost expected accuracy to each observation and to the final answer (Fig.~\ref{fig:visual-example}b).

Applying the audit to five open-weight VLMs and two hosted models reveals a sharp divide between plausible first actions and effective sequential policies. None of the five open models makes a single optimal budget switch in forty pairs each, while GPT-5 mini switches in 48/48 pairs, so the behavior is attainable. In three-read hierarchies, three checkpoints make near-optimal first reads yet lose 22.20--26.76~pp on the routing steps that follow. On natural scanned forms, apparent selection gains move with the reference definition, and two-crop systems stay below native whole-image accuracy while consuming more calls and input tokens.

In summary, we make three contributions:
\begin{itemize}[leftmargin=*,itemsep=2pt,topsep=3pt]
\item \textbf{Budget- and value-paired counterfactual tests} that separate horizon-sensitive planning from fixed-role heuristics under identical public evidence.
\item \textbf{Exact trajectory return accounting} that decomposes end-to-end accuracy loss into stepwise acquisition loss and terminal decision loss.
\item \textbf{An empirical audit across controlled decisions, complete rollouts and natural forms}, showing that later-read routing dominates policy loss and that natural-document gains depend on reference definitions and whole-image comparators.
\end{itemize}

\section{Related Work}
\label{sec:related}
\paragraph{Selective visual observation.}
High-resolution benchmarks and visual search systems study how agents acquire detail beyond an overview~\cite{vstar,hrbench}. FOVEA casts high-resolution vision as sequential Bayesian experimental design, using a coverage--resolution objective to refine crop proposals~\cite{fovea}. Interactive systems such as ReAct, VisualWebArena and SpatialEval further motivate evaluating decisions made across a sequence of observations~\cite{react,vwa,spatialeval}. We examine whether such choices respond to remaining budget and task value, with exact references for controlled cases.

\paragraph{Auditing and crediting visual tools.}
The Illusion of Visual Tool-Use tests reliance on visual returns through policy comparisons, trajectory corruptions and interventions at fixed prefixes~\cite{illusion}. OPO compares alternative observations after shared prefixes to assign local evidence credit during training~\cite{opo}. Our evaluator complements these interventions by assigning each legal alternative a cost in expected terminal accuracy under a declared model. Paired budgets and shifted value boundaries then test behavioral explanations, and return accounting locates the cost across a complete history (Table~\ref{tab:related-distinction}).
\input{related_matrix}

\paragraph{Controlled reasoning tasks.}
CLEVR illustrates how controlled scenes isolate compositional reasoning~\cite{clevr}. AFABench's CUBE-NM includes a context feature that identifies an informative block while providing no immediate label information~\cite{afabench}. We use the same routing principle to construct decisions whose optimal actions differ across matched budgets.

\section{Method}
\label{sec:audit}
The framework combines a declared task model, paired interventions and an evaluator for complete observation histories. Figure~\ref{fig:visual-example} shows the resulting workflow. We first define exact values for each legal observation, then construct pairs that separate candidate first-action rules. Finally, we use those values to account for losses along the tested policy's actual rollout (Algorithm~\ref{alg:audit}).

\subsection{Task Model and Exact Observation Values}
A binding maps each task role to its physical region, and a read model specifies the distribution of the return from a query. In the finite-horizon belief-state formulation~\cite{kaelbling1998planning}, public context $C$ contains the question, rule, overview and verified binding, with scene prior $\pi_C$ and read model $M_C$. A latent scene determines the binary answer; history records acquired observations:
\[
 X\sim\pi_C,\qquad Y=f(X),\qquad H_t=((A_i,Z_i))_{i<t}.
\]
Here $X_a$ is a true scene bit, while $Z_i$ is the possibly noisy return from region $A_i$. Acquisition policy $\mu$ chooses an unrevealed region in $\mathcal A(H_t)$ using $(C,H_t,B-t)$; its randomization is conditionally independent of $X$. The tool appends the return and consumes one observation. After budget $B$ is exhausted, terminal policy $\nu$ predicts the answer. Posterior and channel assumptions are specified in Appendix Sec.~\appref{sec:formal-model}.

The five-region task has selector $S$, branches $T,F$ and two distractors. Under switch, $S$ determines which branch gives the answer; under XOR, the two branch bits determine the label and $S$ is irrelevant. Controlled bit-return tasks specify the channel explicitly, while visual rollouts return actual crops. Verified bindings isolate selection from localization, and the natural-form panel separately evaluates visual choices without assuming an exact scene prior.

Suppressing $C$ for readability, let $p_H=\Pr(Y=1\mid C,H)$. Optimal-continuation values with $b$ reads left are
\begin{align}
 V_0(H)&=\max(p_H,1-p_H),\label{eq:terminal-value}\\
 Q_b(H,a)&=\mathbb E[V_{b-1}(H\oplus(a,Z))\mid C,H,a],\label{eq:action-value}\\
 V_b(H)&=\max_{a\in\mathcal A(H)}Q_b(H,a).\label{eq:optimal-value}
\end{align}
These values give the expected accuracy attainable when action $a$ is followed by optimal continuation, crediting every tied maximizer as an action hit. We define acquisition opportunity loss and model-relative terminal decision loss as
\begin{align}
 \delta_b(H,a)&=V_b(H)-Q_b(H,a),\\
 \eta(H,\hat y)&=V_0(H)-\Pr(Y=\hat y\mid C,H).
\end{align}
Both scores use the declared prior and channel. Reporting $V_0(H)$ alongside $\eta$ distinguishes insufficient evidence from a poor answer given that evidence: at a posterior tie, either answer has zero decision loss, but best-case accuracy is .5. Jointly permuting bindings, histories and the channel preserves all values.

\subsection{Paired Counterfactual Tests}
Fair anchors use independent fair bits and noiseless reads, preserving bindings and legal menus across budgets. A branch is optimal with one read, while the selector is optimal with two. The best suboptimal action loses 25 percentage points (pp) at either budget. Table~\ref{tab:policy-discrimination} shows why the pair is informative: a fixed selector and a horizon-sensitive policy agree at two reads but differ at one. Lowering selector reliability separates a fixed preference from a reliability threshold; shifting the crossing at fixed reliability .74 then tests sensitivity to task value.
\input{policy_discrimination}

Each of forty confirmation families fixes a binding and menu reused across conditions; a profile fixes priors and channel parameters. High-gap anchor parameters are fixed before inference and shared with development. Expressing selected regions as functional roles, we count optimal transitions and measure a separate selector-rate contrast:
\begin{align}
 N_{\rm transition}&=\sum_{i=1}^{40} I_i,\label{eq:transition-count}\\
 I_i&=\mathbf1\{a_i^{(1)}\in\{T,F\},\ a_i^{(2)}=S\},\\
 \Delta_S&=100(\bar s_+-\bar s_-)\quad\text{pp}.\label{eq:adaptation}
\end{align}
The means $\bar s_+,\bar s_-$ compare fixed profiles with positive and negative selector advantage at reliability .74; the exact reference gives 100 pp. Unequal-prior states additionally separate myopic accuracy, entropy reduction and the horizon reference (appendix). These paired queries test stated-budget first choices; sequential execution is evaluated in the hierarchy.

\subsection{Accounting Along the Acquired History}
The finite-horizon performance-difference identity~\cite{trpo} connects the local scores to complete-policy correctness $J_\mu$, with terminal policy $\nu$ fixed. For a legal $B$-read policy under the declared prior and channel,
\begin{equation}
\begin{split}
 V_B(H_0)-J_{\mu}(C,H_0)\\
 =\mathbb E_{\mu,M_C}\!\left[\sum_{t=0}^{B-1}
 \delta_{B-t}(H_t,A_t)+\eta(H_B,\hat Y)\right].
\end{split}
\label{eq:main-return-accounting}
\end{equation}
Successive optimal values telescope by the tower property. The expectation conditions on $(C,H_0)$ and averages latent worlds, read noise and policy randomization (Appendix Sec.~\appref{sec:appendix-accounting}). We report the first loss, the sum over later observations and terminal loss separately. This records how actual policy choices depart from optimal continuation while preserving the histories those choices produce.
\input{audit_algorithm}

\section{Experiments}
\label{sec:experiments}
We test whether observation choices track budget and value, where sequential policies lose return, and how scoring and system baselines change conclusions on natural forms.

\subsection{Experimental Setup}
\label{sec:design}
\paragraph{Models and interfaces.}
We evaluate five open-weight VLMs: Qwen3-VL-4B/8B/32B, InternVL3.5-8B and MiniCPM-V4.5 in fast mode~\cite{qwen3vl,internvl35,minicpmv45}. Core inputs are structured text states with verified bindings. Inference uses BF16 weights, temperature zero and a first-token JSON prefix-tree decoder, with one attempt per query and no explicit reasoning phase. Model components are listed in Appendix Table~\appref{tab:model-components}. Hosted GPT-4.1 and GPT-5 mini runs use their own dated endpoints and effort settings.

\paragraph{Tasks and sample sizes.}
Each model receives 180 acquisition and 108 terminal development requests, followed by 2,320 acquisition and 720 terminal confirmation requests (3,328 in total). Acquisition crosses forty families, eight profiles and seven reliabilities (2,240 queries) and adds eighty budget-paired anchors. Terminal confirmation crosses six fixed histories and three presentations over forty families. Three checkpoints continue to the hierarchy, where forty families and four profiles give 160 three-read episodes per model; its 960 confirmation requests also include isolated one- and two-budget first choices. The FUNSD panel uses six development and thirty confirmation pages from the public training and test splits~\cite{funsd}. Each page supplies a 512-pixel overview and six padded crops in a $3\times2$ grid; pages, questions and both reference definitions are fixed before inference.

\paragraph{Comparators.}
Controlled references include fixed-role policies, one-step accuracy, entropy reduction and the exact remaining-horizon policy. For documents, selection is compared with a shared uniform cached reader, and complete crop systems with native whole-image reading and with overview, blank and GT-located controls.

\paragraph{Evaluation protocol.}
Every model passes a development gate (the complete schedule and at least 95\% strict legality) before confirmation. Action hits accept all tied maximizers; invalid outputs count as misses, and continuous opportunity loss uses legal choices. The core acquisition panels and all hierarchy confirmations are fully legal; document confirmation yields 420/420 legal replies for Qwen3-VL-8B and 418/420 for InternVL3.5-8B (Appendix Table~\appref{tab:new-panel-coverage}). Intervals use 20,000 percentile bootstrap draws over whole families with profiles intact, over terminal components (39 after merging duplicate public inputs), or over whole documents. The four primary endpoints of each model and protocol carry Bonferroni-adjusted 98.75\% intervals; all other contrasts carry 95\% intervals. Bracketed ranges below follow this convention.

\subsection{Budget and Value Sensitivity}
\label{sec:reliability}
All five open-weight VLMs make zero optimal branch-to-selector switches in forty fair pairs each (Table~\ref{tab:open-confirmation}). Qwen3-VL-32B and InternVL3.5-8B choose $S$ on every anchor at both budgets, and Qwen3-VL-8B does so in 39 families; Qwen3-VL-4B and MiniCPM-V4.5 spread their remaining choices over other suboptimal pairs. Correct two-read starts thus coexist with systematic failure at one read.
\input{open_model_main_table}

Mean loss over both anchor budgets is 12.50--15.62~pp, against .59--.85~pp on non-anchor states. Qwen3-VL-32B picks $S$ on all 2,320 confirmation queries, and its .5963~pp non-anchor loss, 12.50~pp anchor loss and zero fixed-reliability contrast reproduce the constant-selector reference exactly. A low near-boundary average hides this fixed preference; the paired anchors expose it.

Budget adaptation is attainable. Under their own protocols, GPT-5 mini makes 48/48 optimal transitions with image plus map and 47/48 with the map alone (the remaining pair starts with a distractor), and GPT-4.1 makes 44/48 with map-only inputs (Table~\ref{tab:hosted-counterexamples}).
\input{hosted_counterexamples_table}

\paragraph{Moving the action-value boundary.}
Four profiles vary the common branch prior and read reliability, moving the selector crossover from .571 to .823. At reliability $\rho=.74$, two profiles favor a branch and two favor the selector (Fig.~\ref{fig:agent-boundary}). Across 24 held-out families, GPT-4.1 selects $S$ on all 864 decisions (480 exact optima), and low-effort GPT-5 mini, despite 567/864 hits, still chooses $S$ in 22--24 families per profile at .74. Predictors fitted on six disjoint development families confirm that reliability, not task value, drives GPT-5 mini: reliability alone gives a held-out selector log loss of .210 nats/action, against .254 for the signed value advantage $D$ (difference .044 [.031, .055]).
\input{review3_figure}

Supplying complete action values fixes the choice. On the 96 GPT-4.1 states at .74, a one-read posterior worksheet gives 44/96 exact hits against 48/96 at baseline, whereas supplying $Q_2(H,a)$ gives 96/96. This 50~pp gain in hits reduces mean opportunity loss by only .5524~pp, because discrete errors near a crossing are cheap; the high-gap anchors are where errors are expensive.

\subsection{Losses Across Complete Trajectories}
\label{sec:extended}
\begin{figure}[t]
\centering\includegraphics[width=.62\linewidth]{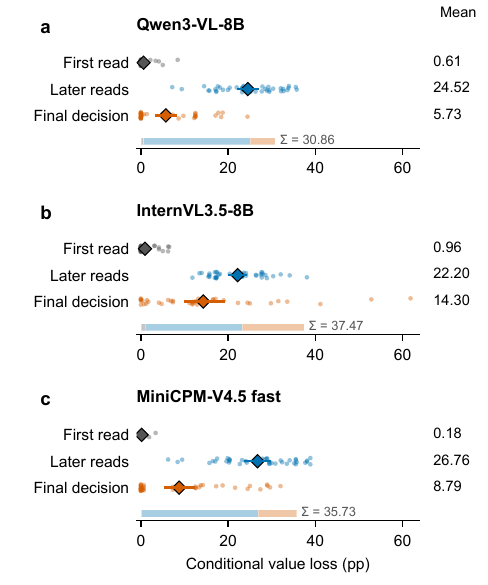}
\caption{Losses across complete three-read histories. Dots show all forty family means over four fixed profiles; diamonds and whiskers give overall means and descriptive 95\% family intervals. Stacked bands accumulate first, later and final decision loss, with totals in pp.}
\label{fig:new-panels}
\par\medskip
\input{hierarchy_primary_table}
\end{figure}

The eight-region hierarchy routes a root through one of two gates to one of four answer leaves, with one distractor. In 160 closed-loop episodes per checkpoint, first loss is .18--.96 pp, while the second and third reads together lose 22.20--26.76 pp (Fig.~\ref{fig:new-panels}; Table~\ref{tab:hierarchy-primary}). Terminal decision loss contributes another 5.73--14.30 pp, and realized accuracy ranges from 53.75\% to 57.50\%. Most of the loss arises after the first observation, when subsequent choices must use the returned bits.

The pattern persists within the balanced noiseless profile: first/later losses are 1.25/35.62 pp for Qwen3-VL-8B, 3.12/29.38 pp for InternVL3.5-8B and 0.00/36.88 pp for MiniCPM-V4.5. In the flat root control, every legal first action ties at .75, yet later loss remains 13.44--14.37 pp. Across profiles, best-case terminal accuracy $V_0(H_B)$ is 60.71--63.98\%; subtracting terminal loss gives declared-model correctness of 56.70\%, 49.67\% and 51.92\% in model order.

Figure~\ref{fig:failure-examples}a follows a recorded Qwen3-VL-8B rollout after an optimal first read. Root W6 returns zero, making gate W4 the routing query, but the policy opens W7 instead and loses 25 pp of conditional value. It then opens W3, leaving two observed leaves in disagreement and the gate unread. The posterior remains tied ($V_0=.5$), so answer zero incurs $\eta=0$ despite disagreeing with the archived truth. The history audit attributes the error to acquisition, which terminal accuracy alone cannot do.
\input{failure_examples}

\subsection{Natural Forms: Selection, Scoring and System Performance}
\label{sec:natural-forms}
For each FUNSD page, one-crop selection chooses among six cached reader outputs for the same question and overview. The uniform comparator averages these exactly. A two-crop policy observes its first return before choosing a second region and answering from both crops; native whole-image reading provides the complete-system comparator.
\input{document_primary_table}
\begin{figure}[t]
\centering\includegraphics[width=.62\linewidth]{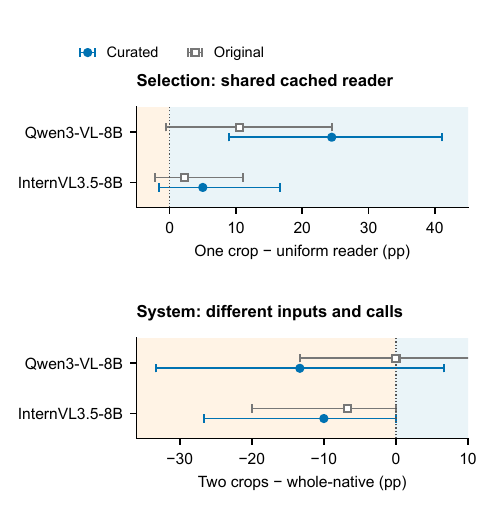}
\caption{Paired contrasts on thirty FUNSD pages. Top: one-crop selection minus the shared uniform reader. Bottom: two-crop system minus native whole-image reading. Circles and squares use curated and original scoring; whiskers give 98.75\% paired-document intervals.}
\label{fig:natural-contrasts}
\end{figure}

Under curated scoring, Qwen3-VL-8B answers 24/30 pages after one-crop selection versus 55.56\% for the uniform reader, gaining 24.44 pp [8.89, 41.11]. Rescoring the same saved decisions against original references gives 10.56 pp [-.56, 24.44], with the interval crossing zero (Table~\ref{tab:document-primary}; Fig.~\ref{fig:natural-contrasts}). InternVL3.5-8B's corresponding gains are 5.00 and 2.22 pp, also with intervals crossing zero. For complete systems, Qwen3-VL-8B answers 21/30 with two crops versus 25/30 natively, a difference of -13.33 pp [-33.33, 6.67]; its overview alone reaches 14/30. InternVL3.5-8B answers 22/30 in each crop condition and 25/30 natively.

Deployment costs sharpen the system comparison. Qwen3-VL-8B's one-crop pipeline uses two calls and 901.2 input tokens per page, while two crops use three calls and 1,662.8 tokens. Native reading uses one call and 824.8 tokens. Their decoded input areas are .561, 1.082 and .766 Mpixel, respectively, counting repeated overviews. With access to all six crops, an oracle reader reaches 86.67\% curated accuracy.

\paragraph{Separating reference text from normalization.}
Both scorers were frozen before inference. A four-cell rescoring crosses original/revised reference text with base/punctuation-space normalization on the same outputs (Table~\ref{tab:document-primary}, Panel B). For Qwen3-VL-8B, text revision alone raises selection gain by 12.22 pp and normalization alone by 10.00 pp; the interaction is -8.33 pp (95\% interval [-17.78, -.56]). Curated scoring changes nine selected-reader and ten native outputs from incorrect to correct, with no reverse flips. Selection accuracy rises 30.00 pp, while the uniform reader rises 16.11 pp (71/180 to 100/180 probes), explaining the 13.89 pp gain shift. Figure~\ref{fig:failure-examples}b shows the Room field, where the original reference merges two form fields.

\subsection{Can State Presentation Repair Terminal Decisions?}
\label{sec:presentation}\label{sec:controller}
To test whether terminal decision loss $\eta$ stems from presenting acquired evidence as prose, we ran a prospective fixed-history intervention that compares prose with a per-region ledger and an active-field control for all five models (720 confirmation queries each; Appendix Sec.~\appref{sec:new-open-models}). The ledger beats the active-field control for Qwen3-VL-4B (+5.83~pp [1.22, 11.54]) and Qwen3-VL-8B (+3.75~pp [.21, 7.50]), yet yields no confirmed gain over prose for any model: +0.42~pp for Qwen3-VL-4B, 0.00~pp for Qwen3-VL-8B, and negative point estimates of $-1.67$ to $-2.92$~pp for Qwen3-VL-32B, InternVL3.5-8B and MiniCPM-V4.5. The development-selected diagnosis also fails to predict where the ledger helps (interaction $-3.00$~pp [$-20.51$, 13.49] for Qwen3-VL-8B). Reformatting public state alone does not repair terminal errors; improvements must target the acquisition and inference policies themselves.

\FloatBarrier
\section{Discussion and Limitations}
\label{sec:implications}
A correct selector choice with two reads can come from a fixed preference for that region. Budget changes separate such a preference from a horizon-sensitive choice, and moving value boundaries tests a finer response to task value. Together with the later losses in complete histories, this argues for conditioning both observation policies and their value critics on the remaining budget and on newly returned evidence. For training, the accounting checks whether credit assigned to an intermediate observation preserves eventual answer value: OPO's local evidence credit~\cite{opo} addresses the learning problem, and our exact controlled cases supply evaluation targets across budgets and subsequent decisions.

Learned observation models would extend the evaluator to less structured scenes, provided their values are accurate. If every estimated action value is within $\epsilon$ of its exact value, choosing the estimated maximizer loses at most $2\epsilon$, and a true gap above $2\epsilon$ preserves the winner (Appendix Sec.~\appref{sec:appendix-accounting}). Near-boundary queries and high-gap anchors therefore play complementary roles in calibrating such models. On natural images, matched-reader selection should be reported beside native whole-image performance and the calls, tokens and pixels each system consumes, and reference definitions should be compared on the same saved outputs.

\paragraph{Limitations.}
The exact evaluator assumes declared priors and read channels, supplied bindings and short read budgets; open scenes require learned observation models whose value errors bound the audit. The five open-weight VLMs share Qwen-derived language components and answer text states under constrained decoding without an explicit reasoning phase, so other model families, free-form reasoning and learned visual channels remain to be tested. Hosted results come from separate protocols with dated endpoints, so they demonstrate attainability rather than a matched comparison. For three of the five models the development rule found no diagnosis-positive state, which leaves the repair-specificity test unidentified for them. The hierarchy is short and its rule is supplied. FUNSD references were curated with AI assistance and no independent human adjudication, and the public pages may have appeared in pretraining data.

\section{Conclusion}
Budget-paired audits show whether observation choices adapt to the reads that remain and where complete histories lose value. In every tested open-weight VLM, later acquisition rather than the first read dominates the loss, and conclusions on scanned forms shift with the scoring reference. Credible evaluation of visual agents requires complete observation histories, matched whole-image baselines and transparent reference scoring.

\FloatBarrier
\bibliographystyle{ieeenat_fullname}
\bibliography{references}

\clearpage
\appendix
\input{appendix_body}

\end{document}

%% file: paper_tables.tex
\usepackage{colortbl}
\definecolor{paperBlue}{HTML}{EAF2F8}
\definecolor{paperAmber}{HTML}{FFF2DB}
\definecolor{paperGray}{HTML}{F2F3F4}

\usepackage{placeins}

%% file: related_matrix.tex
\begin{table}[t]
\centering\small
\setlength{\tabcolsep}{5pt}
\begin{tabular}{@{}lccccc@{}}
\toprule
& Sequential & Counterfactual & Budget/value & Exact per-step & Reference \\
Work & observations & interventions & paired states & loss & scoring audit \\
\midrule
AFABench / CUBE-NM~\cite{afabench} & \checkmark & $\times$ & $\times$ & $\times$ & $\times$ \\
FOVEA~\cite{fovea} & \checkmark & $\times$ & $\times$ & $\times$ & $\times$ \\
Visual tool-use audit~\cite{illusion} & \checkmark & \checkmark & $\times$ & $\times$ & $\times$ \\
OPO~\cite{opo} & \checkmark & \checkmark & $\times$ & $\circ$ & $\times$ \\
\midrule
\textbf{Ours (budget-paired audits)} & \checkmark & \checkmark & \checkmark & \checkmark & \checkmark \\
\bottomrule
\end{tabular}
\caption{Comparison with the closest observation and tool-use evaluations. Prior work studies sequential acquisition, corrupts or replaces returned evidence~\cite{illusion}, or assigns learned local credit to observation actions during training ($\circ$)~\cite{opo}. Our protocol pairs matched budgets and shifted value boundaries with exact finite-horizon loss accounting and an audit of reference scoring on natural forms.}
\label{tab:related-distinction}
\end{table}

%% file: policy_discrimination.tex
\begin{table}[H]
\centering\small
\setlength{\tabcolsep}{2pt}
\begin{tabular}{@{}lccc@{}}
\toprule
Condition & Horizon & Fixed S & Threshold \\
\midrule
Fair, two reads & S & S & S \\
Fair, one read & B & S & S \\
Noisy, reliability .50 & B & S & B \\
Reliability .74, P1--P2 & B & S & S \\
Reliability .74, P3--P4 & S & S & S \\
\bottomrule
\end{tabular}
\caption{Analytical predictions of three first-action rules. B denotes either optimal branch. The illustrative threshold chooses S above reliability .65 and B otherwise, independently of budget and other parameters. All three first choices have value 1.00 at the fair two-read state under optimal continuation. The remaining rows separate their predictions.}
\label{tab:policy-discrimination}
\end{table}

%% file: audit_algorithm.tex
\begin{algorithm}[H]
\caption{Audit one controlled observation episode}
\label{alg:audit}
\small
\begin{algorithmic}[1]
\Require Context $C$, budget $B$, legal policy $\mu$, terminal $\nu$
\State $H\gets H_0$
\For{$b=B,B-1,\ldots,1$}
\State Compute $V_b(H),Q_b(H,\cdot)$ by Eqs.~\eqref{eq:terminal-value}--\eqref{eq:optimal-value}
\State $a\gets\mu(C,H,b)$; record $\delta_b(H,a)$
\State Query $a$, receive $z$, set $H\gets H\oplus(a,z)$
\EndFor
\State $\hat y\gets\nu(C,H)$; record $V_0(H)$ and $\eta(H,\hat y)$
\end{algorithmic}
\end{algorithm}

%% file: open_model_main_table.tex
\begin{table}[t]
\centering\footnotesize
\setlength{\tabcolsep}{2.5pt}
\begin{tabular}{@{}lrrrr@{}}
\toprule
Configuration & \shortstack{$b=1$\\S/B/D} & \shortstack{$b=2$\\S/B/D} & \shortstack{Anchor\\loss} & \shortstack{Other\\loss} \\
\midrule
Qwen3-VL-4B & 26/6/8 & 24/7/9 & 15.62 & 0.85 \\
Qwen3-VL-8B & 40/0/0 & 39/0/1 & 12.81 & 0.64 \\
Qwen3-VL-32B & 40/0/0 & 40/0/0 & 12.50 & 0.60 \\
InternVL3.5-8B & 40/0/0 & 40/0/0 & 12.50 & 0.59 \\
MiniCPM fast & 37/1/2 & 35/3/2 & 13.75 & 0.62 \\
Fixed selector & 40/0/0 & 40/0/0 & 12.50 & 0.60 \\
\bottomrule
\end{tabular}
\caption{Forty paired anchors per configuration. S/B/D counts selector, branch and distractor starts; \textbf{all five models make 0/40 optimal branch-to-selector transitions}. Losses are pp. Full paired categories and 95\% family intervals appear in Tables~\appref{tab:open-budget-pairs} and \appref{tab:anchor-loss-intervals}. Fixed selector is an exact program reference.}
\label{tab:open-confirmation}
\end{table}

%% file: hosted_counterexamples_table.tex
\begin{table}[t]
\centering\footnotesize
\setlength{\tabcolsep}{2.5pt}
\begin{tabular}{@{}llrrr@{}}
\toprule
Model & Input & \shortstack{$b=1$\\S} & \shortstack{$b=2$\\S} & \shortstack{Optimal\\transitions} \\
\midrule
GPT-5 mini & Image + map & 0/48 & 48/48 & 48/48 \\
GPT-5 mini & Map only & 0/48 & 48/48 & 47/48 \\
GPT-4.1 & Image + map & 48/48 & 48/48 & 0/48 \\
GPT-4.1 & Map only & 4/48 & 48/48 & 44/48 \\
\bottomrule
\end{tabular}
\caption{Historical fair-switch decisions in 48 pairs (24 families, two assignments). Optimal transitions require branch-first at $b=1$ and selector-first at $b=2$. GPT-5 mini's map-only exception is a distractor-to-selector pair, showing why selector marginals do not determine paired success. Hosted JSON is parsed after generation; GPT-5 mini uses the endpoint default here and low effort in the boundary panel.}
\label{tab:hosted-counterexamples}
\end{table}

%% file: review3_figure.tex
\begin{figure}[t]
\centering
\includegraphics[width=.62\linewidth]{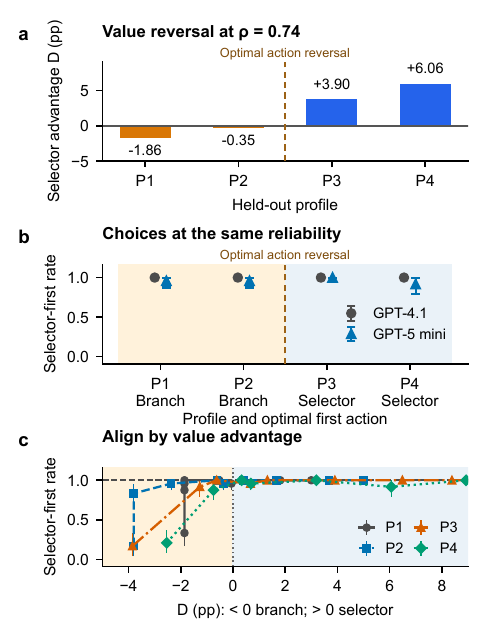}
\caption{Hosted choices at moving boundaries. \textbf{a}, Selector advantage at reliability .74. \textbf{b}, Observed selector rates and optimal roles (24 families/profile; 95\% intervals). \textbf{c}, Choices aligned by signed advantage, accepting all ties at zero. Profiles list (branch prior, read reliability, crossover): P1 (.22,.84,.823), P2 (.30,.80,.754), P3 (.34,.86,.619), P4 (.38,.88,.571). Colored curves show GPT-5 mini at low effort; GPT-4.1 always selects S.}
\label{fig:agent-boundary}
\end{figure}

%% file: hierarchy_primary_table.tex
\centering\footnotesize
\setlength{\tabcolsep}{3pt}
\begin{tabular}{@{}lrrr@{}}
\toprule
Configuration & First / later & Final & Best / actual \\
\midrule
Qwen3-VL-8B & 0.61 / 24.52 & 5.73 & 62.43 / 53.75 \\
InternVL3.5-8B & 0.96 / 22.20 & 14.30 & 63.98 / 57.50 \\
MiniCPM fast & 0.18 / 26.76 & 8.79 & 60.71 / 57.50 \\
\bottomrule
\end{tabular}
\captionof{table}{Three-read histories: 160 per model in forty families. First/later/final are conditional losses (pp); best is $V_0$ and actual is realized accuracy (\%). First-loss and realized-accuracy intervals are 98.75\%; later/final intervals are descriptive 95\% (Appendix Table~\appref{tab:hierarchy-loss-intervals}).}
\label{tab:hierarchy-primary}

%% file: failure_examples.tex
\begin{figure}[t]
\centering\includegraphics[width=.62\linewidth]{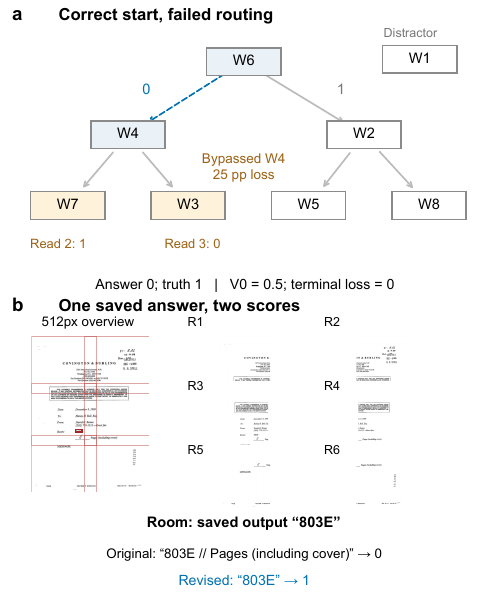}
\caption{Locating acquisition loss and reference sensitivity. \textbf{a}, Recorded Qwen rollout: root W6 returns 0, but the next read bypasses gate W4 and loses 25 pp. The final posterior is tied ($V_0=.5$, $\eta=0$) despite a wrong answer. \textbf{b}, The frozen FUNSD overview and padded crop menu for Room. Removing adjacent-field text from the reference changes the saved answer 803E from exact-match score 0 to 1.}
\label{fig:failure-examples}
\end{figure}

%% file: document_primary_table.tex
\begin{table}[t]
\centering\small
\setlength{\tabcolsep}{3pt}
\begin{tabular}{@{}lrr@{}}
\toprule
\multicolumn{3}{@{}l}{\textbf{Panel A: Paired document contrasts}} \\
Reference & \shortstack{One crop $-$\\uniform (pp)} & \shortstack{Two crops $-$\\native (pp)} \\
\midrule
\multicolumn{3}{l}{\textbf{Qwen3-VL-8B}} \\
Curated & \cellcolor{paperBlue}\shortstack{24.44\\{}[8.89, 41.11]} & \shortstack{-13.33\\{}[-33.33, 6.67]} \\
Original & \cellcolor{paperBlue}\shortstack{10.56\\{}[-0.56, 24.44]} & \shortstack{0.00\\{}[-13.33, 13.33]} \\
\multicolumn{3}{l}{\textbf{InternVL3.5-8B}} \\
Curated & \shortstack{5.00\\{}[-1.67, 16.67]} & \shortstack{-10.00\\{}[-26.67, 0.00]} \\
Original & \shortstack{2.22\\{}[-2.22, 11.11]} & \shortstack{-6.67\\{}[-20.00, 0.00]} \\
\bottomrule
\end{tabular}
\par\medskip
\textbf{Panel B: Selection gain under four scorers (pp)}\par\smallskip
\begin{tabular}{@{}llrr@{}}
\toprule
Reference text & Normalizer & Qwen & InternVL \\
\midrule
Original & Base & 10.56 & 2.22 \\
Revised & Base & 22.78 & 5.00 \\
Original & Punctuation & 20.56 & 2.22 \\
Revised & Punctuation & 24.44 & 5.00 \\
\bottomrule
\end{tabular}
\caption{Thirty paired FUNSD pages. \textbf{A}, One-crop selection minus shared uniform reading and two crops minus native reading, with 98.75\% paired-document intervals. Shading pairs Qwen's scoring definitions. \textbf{B}, Original/revised reference text crossed with base/punctuation-space normalization on the same saved outputs. Full four-cell intervals and oracle results appear in Table~\appref{tab:scoring-four-cell-full}.}
\label{tab:document-primary}\label{tab:scoring-factorial}
\end{table}

%% file: appendix_body.tex
\setcounter{topnumber}{2}
\setcounter{section}{0}
\setcounter{table}{0}
\setcounter{figure}{0}
\renewcommand{\thesection}{A\arabic{section}}
\renewcommand{\thetable}{A\arabic{table}}
\renewcommand{\thefigure}{A\arabic{figure}}
\let\AppendixSection\section
\renewcommand{\section}{\FloatBarrier\AppendixSection}
\paragraph{Organization.} This appendix follows the order of the evidence in the main paper. Sections~\ref{sec:formal-model}--\ref{sec:protocol-summary} define the task model and summarize every protocol. Sections~\ref{sec:stimuli}--\ref{sec:matched-replay} document the historical controlled panels, exact values, scoring rules and matched-world controller replays. Sections~\ref{sec:supp-v2}--\ref{sec:matched-downstream} cover the confirmation stimuli, public-state presentation and the moving selector boundary. Section~\ref{sec:new-open-models} reports the matched-protocol open-model study behind the paired-anchor results, and Section~\ref{sec:extended-methods} the three-read hierarchy and FUNSD panels. Section~\ref{sec:appendix-accounting} proves the return-accounting identity, and Section~\ref{sec:scope} states the scope of the evidence.

\input{formal_model_details}
\input{agent_design_table}
\input{agent_protocol_summary}

\section{Stimulus Grammar and Construction}\label{sec:stimuli}
The active region vocabulary is $W_1,\ldots,W_5$. Region IDs specify physical locations and are the legal response tokens while a reveal remains. The visible labels A, B, C, X and Y are distinct symbols, each placed above one region. A, B and C occupy three task-relevant slots. X and Y occupy two distractor slots. The displayed rule maps A, B and C to functional roles $S,T,F$. Those roles are defined by the rule, while the physical IDs and visible labels belong to the image. In the switch rule, $S$ selects the answer source, $T$ is its branch for $S=1$, and $F$ is its branch for $S=0$. Under branch XOR the nominal $S$ is irrelevant and the output is $X_T\oplus X_F$. A family groups matched bindings, layouts or latent worlds for pairing and resampling, as specified for each panel. In this base visual task, it shares an image, binding and sampled closed-loop world across interventions.

Stimuli were generated deterministically. For family index $i=0,\ldots,9$, a Python pseudorandom generator with seed 2026111007 shuffles the ordered tags (A,B,C). The task tags are assigned to slots with zero-based indices $(i+d)\bmod 5$. The offsets $d$ are $(0,1,3)$ for the first five families and $(0,2,4)$ for the next five. The leftover slots receive X and Y in randomly sampled order. The two cycles vary the spacing between relevant windows, with each ordinal role visiting every physical position once per cycle. Each $700\times240$ white image renders five rounded rectangles at $x_j=16+139j$ and $y_j=78+\epsilon_j$, with $\epsilon_j$ an integer in $[0,23]$. Each rectangle is $110\times85$ pixels in drawing coordinates, with radius 12 and a three-pixel outline. Tags and physical IDs use Arial Bold at 25 pixels. Each hidden bit is sampled as a fair Bernoulli draw for the base closed-loop world after writing the image. The image contains only IDs, tags and outlines. The manifest records the seed and SHA-256 hashes of the script, plan, font and rendered assets. Image hashes are checked against prior experimental images at generation and against the frozen manifest at replay.

\begin{figure}[!htbp]
\centering
\includegraphics[width=.6\linewidth]{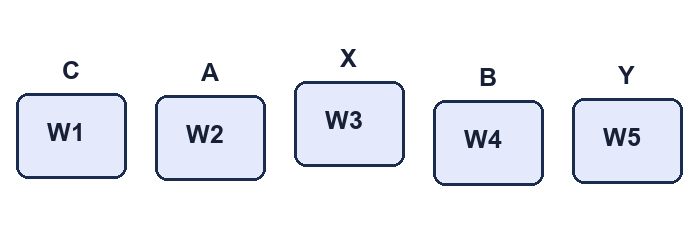}\\[3pt]
\includegraphics[width=.6\linewidth]{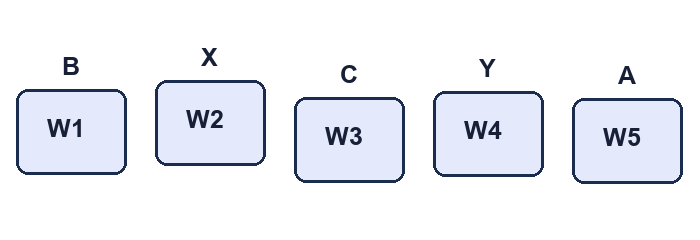}
\caption{Actual held-out renderings from the two spacing cycles. The private worlds and role bindings are stored in the manifest.}
\label{fig:stimuli}
\end{figure}

A worked example from the upper image in Figure~\ref{fig:stimuli} uses $S=\mathrm{C}=W_1$, $T=\mathrm{A}=W_2$ and $F=\mathrm{B}=W_4$. The rule says to use A if C is 1 and B otherwise. Revealing $W_1=1$ makes $W_2$ the sole relevant branch, while the original image stays byte-identical. The actual frozen world has $W_2=0$ and $W_4=1$. The rule's answer is 0.

The primary image payload contains the PNG, an explicit tag-to-ID map, the rule, the priors, the ordered reveal history, the remaining count and the legal codes. Previously revealed windows are removed from the menu. At zero reveals, the legal responses are 0 and 1. The image-only binding probe omits the map and asks for A, B and C by physical ID. The fully observed rule probe supplies all rule operands. The structured diagnostic omits the PNG and preserves the map and instructions. The forced-history panel intervenes on the visible selector bit and branch ordering at one remaining reveal. Unequal-prior stimuli supply declared priors to first-action decisions. The closed-loop outcomes use the fair base worlds.

\section{Experimental Panels and Artifacts}
\begin{table}[!htbp]
\centering\small
\begin{tabular}{@{}p{.24\columnwidth}lp{.52\columnwidth}@{}}
\toprule
Evidence source & Families & Experimental conditions\\
\midrule
Historical Sonnet closed loop & 30 & Two budgets on the same families. Retains selection order, visible values, answer and ground truth.\\
Historical Sonnet paired actions & 24 & Rule, selector assignment, budget and forced history in a separate archive.\\
Historical Qwen3.5-4B holdout & 24 & One shared map per family, 72 independent rule checks and 192 direct decisions. Greedy and horizon controllers replay the 192 states.\\
Frozen Sonnet decision panel & 10 & Eight fair starts per family (2 assignments $\times$ 2 rules $\times$ 2 budgets), two unequal-prior image starts, one map-only start at budget two, four forced-history decisions, one binding probe and two rule probes.\\
Forced-tool sensitivity & Same 10 & One first-action request per image and budget for the 20 unequal-prior image cells. Sessions remain separate.\\
\bottomrule
\end{tabular}
\caption{Historical and initial frozen Sonnet evidence sources. The two assignments per fair rule and budget yield 20 cells nested in ten families. Later protocols are summarized separately in Table~\ref{tab:protocol-summary}.}
\label{tab:panels}
\end{table}

The frozen decision matrix in Table~\ref{tab:panels} totals 180 planned first requests, including 150 scored decision conditions. The first calls for the base fair switch under both budgets also start closed-loop traces on the same private world. The 24 additional calls are continuations and final answers after a valid preceding action. First-call order is randomized with seed 2026111008. Each request starts from a fresh context with the public evidence needed for the current step. Wording uses the parity of the family index and flips across assignments in the fair panel.

The forced-tool sensitivity uses the same 20 unequal-prior image decisions, with the instruction and response contract changed to a single structured observation-selection call. It preserves the map, rule, priors, menu, budgets and image bytes. Request order was fixed with seed 2026111013. The first request doubled as the capability check and succeeded. The fallback was unused. No trial was retried. The tool returns one legal window in one field and the trial ends at first choice. Two timeouts and two TLS disconnects left missing responses.

\begin{table}[!htbp]
\centering\small
\begin{tabular}{cll}
\toprule
Stimulus & Labels at $W_1,\ldots,W_5$ & Base roles $(S,T,F)$ \\
\midrule
1 & C, A, X, B, Y & C, A, B \\
2 & Y, B, A, X, C & B, A, C \\
3 & B, Y, C, A, X & C, A, B \\
4 & Y, A, X, B, C & B, C, A \\
5 & B, X, A, Y, C & C, B, A \\
6 & B, X, C, Y, A & B, C, A \\
7 & A, C, X, B, Y & C, B, A \\
8 & X, C, A, Y, B & A, B, C \\
9 & C, X, B, A, Y & A, C, B \\
10 & X, C, Y, B, A & A, C, B \\
\bottomrule
\end{tabular}
\caption{Role assignments in the ten confirmatory stimuli. The second fair-rule assignment exchanges $S$ and $T$ while retaining the same image and physical bindings.}
\label{tab:families}
\end{table}

\section{Normative Policies and Exact Values}
This subsection specializes the general model to the base noiseless task with independent bits. Define $p_a(H)=\Pr(X_a=1\mid C,H)$: an observed bit fixes this probability to its returned value, while an unobserved bit retains its prior. Under switch, $p_H=p_S(H)p_T(H)+[1-p_S(H)]p_F(H)$; under XOR, $p_H=p_T(H)[1-p_F(H)]+[1-p_T(H)]p_F(H)$. Independence is preserved here by separate noiseless reads. In noisy tasks one must update the latent-bit posterior using the read channel rather than equating a noisy return with the true bit.

The final Bayes accuracy is $V_0(H)=\max(p_H,1-p_H)$. For a legal region $w$,
\begin{align*}
 Q_b(H,w)&=\sum_{z\in\{0,1\}}\Pr(Z=z\mid C,H,w)\\
 &\qquad{}\times V_{b-1}(H\oplus(w,z)),\\
 V_b(H)&=\max_{w\in\mathcal A(H)}Q_b(H,w).
\end{align*}
One-step accuracy ranks $Q_1$ even if $b>1$; horizon accuracy ranks $Q_b$. Entropy reduction ranks $h(p_H)-\mathbb E[h(p_{H\oplus(w,Z)})\mid C,H,w]$, where $h$ is binary entropy in bits. Baselines include the first legal physical location and, when unrevealed, the nominal selector. Hits use the full argmax set. Deterministic replay breaks action ties by menu order and terminal ties by choosing 0.

\begin{table}[!htbp]
\centering
\small
\setlength{\tabcolsep}{3pt}
\begin{tabular}{@{}p{.42\columnwidth}rrrr@{}}
\toprule
First action & $S$ & $T$ & $F$ & \shortstack[r]{Either\\distractor}\\
\midrule
Fair switch $Q_1$ & .50 & \textbf{.75} & \textbf{.75} & .50\\
Fair switch $Q_2$ & \textbf{1.00} & .75 & .75 & .75\\
Fair XOR $Q_1$ & \textbf{.50} & \textbf{.50} & \textbf{.50} & \textbf{.50}\\
Fair XOR $Q_2$ & .50 & \textbf{1.00} & \textbf{1.00} & .50\\
Unequal-prior entropy reduction (bits) & .04616 & \textbf{.15225} & .11811 & 0\\
Unequal-prior $Q_1$ & .820 & .820 & \textbf{.864} & .820\\
Unequal-prior $Q_2$ & \textbf{1.000} & .954 & .880 & .864\\
\bottomrule
\end{tabular}
\caption{Exact first-action values at empty history. The unequal-prior switch uses priors .4, .3 and .1 for S, T and F, with .5 for distractors. Bold indicates the optimum.}
\label{tab:q}
\end{table}

Table~\ref{tab:q} separates conditional value from realized performance. With fair bits, observing either branch supplies its answer with probability $\frac12+\frac12\frac12=.75$. Seeing the selector alone identifies a branch whose value is still fair. With two reveals, selector then active branch determines the output. A distractor followed by a branch is worth .75. When both revealed branches disagree and the selector remains hidden, $p_H=.5$. The archive's two correct answers at that posterior are realized hits under residual uncertainty.

The unequal-prior baseline is $p_H=.4(.3)+.6(.1)=.18$. With one reveal, horizon and greedy accuracy both select $F$. All three predictions differ at two reveals. The respective best-minus-runner-up gaps are .03414 bits for entropy, .044 expected accuracy for one step, and .046 expected accuracy for two steps. The nominal selector also predicts the two-step optimum on these states, so a two-step hit alone does not identify planning. In the forced-history panel, a revealed selector deterministically specifies one branch and revealing it attains 1.00.

\section{Scoring and Completeness}
\textbf{Endpoints.} For valid actions, report exact target hits and $V_b(H)-Q_b(H,w)$. The fair budget shift codes selector choices as 1 and other choices as 0, and averages two assignment-specific differences within each family before comparing switch and XOR. Forced continuation scores isolated hits and complete counterfactual flips. A completed binary answer is scored against the archived private world and active rule. Equal terminal correctness on a small set of realized worlds is compatible with different normative action values.

\textbf{Response handling.} The original confirmation requires an exact legal code after whitespace and case normalization. A completed but invalid reply is unsuccessful for the scheduled hit rate and zero for selector choice, with no action regret. Transport failures are missing with no behavioral score. The exploratory parser uses an explicit legal action on the last line of a normally terminated reply and remains separate from the primary analysis. The forced-tool parser accepts one named tool-use block containing exactly one legal window field. Closed-loop continuation stops after an invalid or missing action.

\textbf{Family dependence.} Family bootstrap intervals use 10,000 percentile draws (seed 113 for confirmation and forced-tool comparisons). The historical 30-family bootstrap uses seed 20261110. These intervals describe the frozen families and collapse at zero variation. The fair-panel shift uses ten families, preserving their shared assignments and rules. The forced-tool comparison uses six complete pairs and separately bounds the scheduled-cell shift across all $2^4$ completions of the missing selector indicators. The bound is $+30$ to $+70$ percentage points.

\textbf{Hosted Sonnet.} The archived system was Sonnet 4.6, requested temperature 0 and a requested maximum of 256 tokens. The 204 original calls produced 32 invalid first replies across the 150 decision conditions and no transport failures. Recorded usage was 16,395 uncached input and 107,822 output tokens, with 743.94 summed wall-clock seconds across concurrent requests. Some reported output counts exceeded the requested cap. These measurements are gateway-reported, and summed request times differ from elapsed time. The 20 forced-tool requests were a separate session under the same routed identifier.

\textbf{Open-weight Qwen.} The historical archive uses the MLX 4-bit conversion of Qwen3.5-4B at a fixed revision recorded with the artifact, an Apple M3 with 8 GiB unified memory, MLX 0.32.4 and MLX-VLM 0.7.12. Generation requested temperature zero, disabled thinking and capped output at 256 tokens. The eight-family development split used 32 screening requests. The separate 24-family holdout used 24 mapping calls, 72 rule checks and 192 direct decisions. The controllers replay the same 192 action states with those maps, with no additional model calls. They differ in search horizon. Offline scoring separates infrastructure attempts from behavior.

\section{Unequal-Prior Session Comparison}
\paragraph{Choices change under unequal priors, with response-format sensitivity.}
Free text yielded five legal actions across 20 image trials, all $S$. Dropping the image at budget two yielded eight legal actions across ten trials, seven at $S$. A separate forced-tool session used the same images, maps, budgets and served model identifier, and returned 16 legal actions with four missing in transport. At budget one, the eight legal choices were three $S$, four $T$ and one $F$. At budget two, all eight legal choices were $S$. Among six complete pairs, four changed from $T$ to $S$ and two stayed at $S$ (Table~\ref{tab:format-sensitivity}). The sole $F$ choice had no observed partner. The paired selector shift was $+66.7$ pp ($[33.3,100]$), with a scheduled-cell bound of $+30$ to $+70$ pp over possible missing choices. Thus the forced-tool observations show budget sensitivity on this panel, but the complete pairs contain no instance of the specific horizon prediction, a change from $F$ to $S$. The separate sessions leave the cause of the format-associated difference unresolved.

\begin{table}[!htbp]
\centering
\setlength{\tabcolsep}{2pt}
\begin{tabular}{@{}lrrrrrr@{}}
\toprule
Condition & $n$ & $S$ & $T$ & $F$ & Inv. & Miss. \\
\midrule
Free text, both &20&5&0&0&15&0\\
Tool, $b=1$ &10&3&4&1&0&2\\
Tool, $b=2$ &10&8&0&0&0&2\\
\bottomrule
\end{tabular}
\caption{Unequal-prior first actions. The tool session retained six complete budget pairs; four changed from $T$ to $S$ and two remained at $S$. The only observed $F$ action had no partner. Separate sessions and incomplete pairs limit attribution to response format.}
\label{tab:format-sensitivity}
\end{table}

\section{Reproducibility and Validation}\label{sec:replay}
The accompanying artifact provides the frozen stimuli, random seeds, prompts, anonymized responses and analysis programs. Checksums preserve the link between each observation and its source. Independent validation enumerates all $2^5$ latent worlds, checks 150 exact-action conditions, verifies that changing hidden values leaves public inputs unchanged, and checks the treatment of transport failures.

\section{Trajectory Verification and Observation-Noise Analysis}
An offline replay reconstructs 60 archival episodes from the historical reanalysis. It verifies first and continuation values and records the posterior support for the selected answer. Among the 29 completed two-reveal episodes, 12 ended at a posterior tie (7 correct), one ended with partial evidence (1 correct), and 16 ended with decisive evidence (15 correct). In two trajectories, an optimal selector start was followed by an inactive-branch read, losing .50 in conditional action value. In another trajectory, both reads had zero regret and the answer disagreed with decisive evidence.

The same script performs an exact reliability extension with the switch rule, independent latent priors $(.50,.30,.30)$ and branch read reliability .86. Each read is a conditionally independent symmetric bit-flip observation. The selector reliability ranges from .50 to 1 in .01 increments, with the five legal region IDs and costs fixed. The answer uses latent bits rather than noisy observations. The factorized Bayesian update and an independent 32-world recursion agree on all 102 reliability-by-budget cells. Near the crossover, the selector value is $Q_2(S)=.5576+.3024r_S$ and each branch has value $Q_2(T)=Q_2(F)=.75696$. Equating these expressions gives $r_S=89/135$. We checked this tie with both implementations using exact rational arithmetic. The complete exact values and optimal-action sets accompany the artifact. The test characterizes an explicit observation model. This offline calculation uses no model responses.

\section{Reconstructing the Public Agent State}
The step-level archive audit reads 60 saved Sonnet episodes and every corresponding prompt and reply. It checks the stimulus hashes, recovers the region IDs, labels and values serialized by the observation tool, and verifies the legal menu and exact values at every acquisition. It compares terminal answer support with the posterior mode. For the archived visual policy the full binding was not a requested output. The audit verifies 30 and 87 serialized label instances at budgets one and two, respectively, without treating that check as a model grounding score. A separate binding probe can measure the latter quantity.

One-reveal terminal histories were tied in six episodes (three correct) and had .75 mode support in 24 (18 correct). The model gave a nonmodal answer in one trajectory. Among the 29 completed two-reveal trajectories, 12 were tied (seven correct), one had .75 support (one correct) and 16 were decisive (15 correct). The two positive continuation losses occurred in two trajectories, each an inactive-branch reveal after an optimal selector start. The only nonmodal two-reveal answer followed two optimal actions that had already identified the answer. The incomplete two-reveal episode failed at the first action. All 60 public histories are retained in the artifact.

\section{Additional Decision and Controller Evidence}
\begin{figure}[!htbp]
\centering
\includegraphics[width=.6\linewidth]{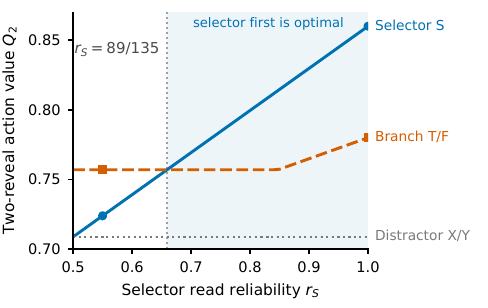}
\caption{Exact two-reveal action values $Q_2$ when only the selector read is noisy. Latent priors are .50 for $S$ and .30 for each branch; branch and distractor reads are correct with probability .86. Inspecting the selector first becomes optimal above $r_S=89/135\approx.659$ (shaded). Markers show the two worked reliabilities, .55 and 1. The two branches share one curve, as do the two distractors.}
\label{fig:offline-reliability}
\end{figure}

Table~\ref{tab:confirmation} reports fair-prior selector starts. Figure~\ref{fig:offline-reliability} gives exact action values under simulated observation noise. The reference controller updates the evidence state and remaining budget after each actual observation; candidate observations during search remain hypothetical.
\begin{table}[!htbp]
\centering
\small
\setlength{\tabcolsep}{3pt}
\begin{tabular}{@{}lccp{.40\linewidth}@{}}
\toprule
Rule & $b=1$ & $b=2$ & Reference \\
\midrule
Switch & \cellcolor{paperAmber}19/20 & \cellcolor{paperBlue}19/20 & Selector is suboptimal with one reveal and optimal with two. \\
Branch XOR & 0/20 & 0/20 & Selector is irrelevant at either budget. \\
\bottomrule
\end{tabular}
\caption{Sonnet 4.6 selector starts with the same pixels across budgets. Each denominator comprises two role assignments in each of ten families. Shading distinguishes suboptimal and optimal selector choices; counts within a family are repeated measurements.}
\label{tab:confirmation}
\end{table}

\section{Matched-World Replay of the Reference Controllers}\label{sec:matched-replay}
We replay the two reference policies on the 30 private worlds from the archived Sonnet closed-loop panel. The replay verifies that all recorded native observations match those worlds and that the archived answer key matches the rule. The controller receives the verified map, fair priors, rule, budget and revealed history. The archive returns each hidden bit only when the policy selects its window. Both controllers use the same exact value code and choose the terminal posterior mode. Action ties resolve by window order W1--W5, and answer ties resolve to 0. At budget two, the policies can choose different first windows, so their subsequent histories need not coincide. Table~\ref{tab:archive-replay} preserves the original interface and scoring for native Sonnet, with the invalid two-reveal start counted as a miss.

\begin{table}[!htbp]
\centering
\small
\begin{tabular}{@{}lp{.42\columnwidth}rr@{}}
\toprule
Budget & Policy & \shortstack[r]{Optimal first\\actions} & \shortstack[r]{Correct\\answers}\\
\midrule
1 & Native Sonnet 4.6 & 24/30 & 21/30 \\
1 & One-step reference & 30/30 & 22/30 \\
1 & Remaining-budget reference & 30/30 & 22/30 \\
2 & Native Sonnet 4.6 & 9/30 & 23/30 \\
2 & One-step reference & 0/30 & 21/30 \\
2 & Remaining-budget reference & 30/30 & 30/30 \\
\bottomrule
\end{tabular}
\caption{Native and replayed outcomes on the same 30 latent worlds per budget. The difference between the reference policies isolates lookahead with matched grounding, observation budget and terminal inference.}
\label{tab:archive-replay}
\end{table}

\input{confirm_v2_appendix}

\IfFileExists{agent_factorial_appendix.tex}{%
\section{Factorial Public-State Presentation}
\input{agent_factorial_appendix}
\input{review3_distribution_table}
}{}

\section{Moving the Selector Decision Boundary}
The selector prior stays .5. The branch priors are both $p$ and the branch and distractor reads have symmetric correctness $r_B$. The selector read has correctness $\rho$. We enumerate the two-reveal policy values. A dense grid checks that each profile has one crossing, followed by bisection for the displayed threshold. The held-out profiles differ in both priors and read reliability (Table~\ref{tab:boundary-profiles}). For each, the query grid includes seven common reliabilities from .50 to 1.00 and two extra points close to the threshold. Every one of 24 held-out families receives all 36 settings. Region binding and legal-menu order are fixed within each family. Strict JSON accepts one legal region ID per request.

\begin{table}[!htbp]
\centering
\begin{tabular}{@{}rrrr@{}}
\toprule
$p$ & $r_B$ & Exact crossing & $D$ at $\rho=.74$ \\
\midrule
.22 & .84 & .8225 & $-.01858$ \\
.30 & .80 & .7540 & $-.00352$ \\
.34 & .86 & .6192 & $+.03902$ \\
.38 & .88 & .5708 & $+.06060$ \\
\bottomrule
\end{tabular}
\caption{The same selector reliability can imply opposite first actions. $D$ is the expected accuracy advantage of inspecting the selector over the best alternative.}
\label{tab:boundary-profiles}
\end{table}

Three regularized logistic fits predict the binary selector choice from reliability alone, from $D$ alone, or from reliability, $p$ and $r_B$. They fit 216 decisions from six development families and four other profiles, then score the 864 decisions on the held-out families and profiles. For a separate action-level comparison, horizon, one-step accuracy, one-step entropy and selector-role policies mix a uniform distribution over their own best actions with a fitted uniform lapse over the legal menu. The lapse is fitted on development choices. Both comparisons score held-out log loss (Table~\ref{tab:boundary-confirmation}). GPT-5 mini uses low reasoning effort in this boundary protocol. That setting was selected on development after checking output truncation. The separate factorial comparison uses the endpoint default. Their predictions characterize behavior under the tested inputs.

\input{agent_boundary_appendix}

\paragraph{Computed-state follow-up at a common reliability.}
All four profiles at $\rho=.74$ contribute 24 states each. We preserve the original GPT-4.1 action menu, rule, map and strict enum. The one-read worksheet adds $\Pr(Z=1\mid C,H,a)$ and $\Pr(Y=1\mid C,H,a,Z=z)$ for both returns at every legal region. The two-step worksheet instead adds $Q_2(H,a)$ for every region, in menu order. No optimum is labeled. Both aids add external computations to the same public facts. This follow-up reuses the evaluated states after the initial diagnosis.

The baseline had 48/96 optimal actions. The one-read worksheet had 44/96, with a paired change of $-4.17$ pp ($[-11.46,4.17]$), and increased action loss by 1.41 pp ($[1.01,1.74]$). The two-step worksheet had 96/96, with a paired change of $+50$ pp and an action-loss reduction of .5524 pp. The latter gains occurred in the two profiles where $\rho=.74$ is below the true crossing. Because the worksheet supplies computed action values, this result measures how the model uses that reference rather than whether it can compute it.

\section{Controlled State-Presentation Evidence}
\input{agent_presentation}
\input{legacy_intervention_details}

\paragraph{A supplied route and a chosen route differ.}
On 20 forced-tool Sonnet trials, a step prompt's accidental ``window 2'' string overrode the intended first response. Later trials followed the supplied selector value in 31/40 scheduled cases, while native fair starts chose the selector at both budgets in 19/20 cases each. These historical observations concern continuation given a route, rather than successful selection of that route.

\section{Matched Downstream State Presentation}\label{sec:matched-downstream}
The GPT-4.1 rollout evaluates 24 previously used masked confirmation families under both switch and XOR. Each joint episode shares the first VLM action and the read of its first crop. The two arms use the same legal menu and strict JSON enum to select a second region, then use the same reader for each requested crop. The recorded history contains that reader's actual output. A read is reused if both policies choose the same second region. At termination the prompt keeps the same rule, map and crop images, with the history expressed as prose or as one row per region. An arm can see only the crops acquired on its own path. This isolates a downstream change in representation, conditional on the common first decision.

All 48 joint episodes completed. They used 338 scored model steps and 348 requests including retries. The audit made 98 distinct crop reads, all correct. The switch arms had 23/24 and 24/24 optimal continuations for prose and ledger, respectively; the XOR arms had 24/24 and 23/24. The switch answer scores were 21/24 and 23/24, and XOR scores were 22/24 and 23/24. The paired 95\% interval for the pooled difference in continuation hits was $[-6.25,6.25]$ pp. For the 46 same-second-region pairs the answer totals were 41/46 and 44/46. All three improvements had decisive observed evidence and a nonmodal prose answer. We describe this subset after observing the trajectories. The primary contrast uses all 48 pairs and resamples the 24 families with their two rules intact.

\input{open_model_methods}
\input{new_panel_methods}
\input{diagnostic_accounting}

\section{Scope and Extension}\label{sec:scope}
The task measures discrete choices under explicit binary observations, short fixed budgets and supplied rules. It has no eye movement, stopping-time, human participant or same-task human comparison data. Supplied maps isolate conditional choice in the primary panel. The mapping probes and structured diagnostic change the prompt and measure different operations. The deterministic controller contrast identifies the effect of horizon under a fixed formal model and shared map. The direct model can differ from a controller in inference, representation and output constraints.

To apply the controller to natural-image crops, estimate candidate observation distributions after grounding regions, update evidence with perceptual uncertainty, and evaluate action values at the remaining budget. Exact enumeration assumes declared Bernoulli probabilities and either noiseless or specified symmetric-error read models. The reliability decisions test stated channel parameters; the reported crop rollouts use readable noiseless fields. It leaves the quality of learned observation models for natural images to empirical testing. Hosted-version dependence, a limited family panel and missing tool responses constrain generalization. An evaluation that isolates visual acquisition should cross supplied and inferred bindings, compare text and crop returns, and hold the candidate menu and reveal budget fixed when changing the controller.

\FloatBarrier

%% file: formal_model_details.tex
\section{Notation, Posterior and Decision Model}
\label{sec:formal-model}
Table~\ref{tab:notation} collects the controlled-task notation. The context $C$ includes all initially available information. In map-only tasks, $\pi_C$ is the supplied prior after conditioning on that context; in controlled pixel tasks the overview shows layout without revealing hidden bits. The natural-form panel does not posit such a prior and therefore reports empirical correctness, not these exact values.

\begin{table}[!htbp]
\centering\small
\begin{tabular}{@{}lp{.70\columnwidth}@{}}
\toprule
Symbol & Meaning \\
\midrule
$C,G,f$ & Fixed public context, role binding and answer rule. \\
$X,\pi_C$ & Latent scene and its conditional prior. \\
$X_a,Y$ & True region bit and answer $Y=f(X)$. \\
$p_a,r_a$ & True-one prior and read correctness for region $a$. \\
$A_t,Z_t$ & Region selected at step $t$ and tool return. \\
$M_C$ & Declared conditional observation kernel. \\
$H_t$ & Ordered acquired history before read $t$. \\
$\mathcal A(H)$ & Unrevealed legal regions at history $H$. \\
$B,b$ & Initial and remaining observation budgets. \\
$\mu,\nu$ & Acquisition and terminal-answer policies. \\
$p_H$ & Posterior $\Pr(Y=1\mid C,H)$. \\
$V_b,Q_b$ & Optimal terminal return and action value. \\
$\delta_b,\eta$ & Acquisition loss and model-relative terminal decision loss. \\
$J_\mu$ & Expected correctness of the complete policy. \\
$D$ & Selector value minus the best alternative. \\
\bottomrule
\end{tabular}
\caption{Symbols for exact controlled-task diagnostics. $\pi_C$ is a prior, never a policy; $Z_t$ is an observation, not the latent bit $X_{A_t}$.}
\label{tab:notation}
\end{table}

\paragraph{Assumptions.}
The latent scene is fixed during an episode. Each action uses only $(C,H_t)$ and randomization independent of $X$ conditional on that information; no hidden scene value enters its prompt. The terminal policy $\nu(\hat y\mid C,H_B)$ has the same information restriction. All scored histories have positive probability under the declared prior and observation kernel. Actions are legal, with enough unrevealed regions to exhaust the fixed budget. The prior and kernel must describe the evaluation environment for the return interpretation to hold. Independence of bits and symmetric read noise are properties of particular controlled tasks, not necessary for the accounting identity.

\paragraph{Posterior and terminal policy.}
For a finite latent scene and a realized history $h_t=((a_i,z_i))_{i<t}$, Bayes' rule gives
\begin{equation}
 \Pr(X=x\mid C,h_t)
 =\frac{\pi_C(x)\prod_{i<t}M_C(z_i\mid x,a_i,h_i)}
 {\sum_{x'}\pi_C(x')\prod_{i<t}M_C(z_i\mid x',a_i,h_i)}.
 \label{eq:task-posterior}
\end{equation}
The action-selection factors $\mu(a_i\mid C,h_i)$ cancel because they do not depend on $x$ after conditioning on the public history. If actions use extra private observations of $X$, this cancellation and the evaluator's posterior need not be valid. The predictive observation distribution is the posterior mixture of $M_C(\cdot\mid x,a,H)$. Summing the posterior over $\{x:f(x)=1\}$ yields $p_H$. The resulting recursion is $V_0=\max(p_H,1-p_H)$, $Q_b=\mathbb E[V_{b-1}(H\oplus(a,Z))\mid C,H,a]$ and $V_b=\max_a Q_b$.

For stochastic terminal decoding, its conditional correctness is
\begin{equation}
 R_\nu(H)=\sum_{y\in\{0,1\}}\nu(y\mid C,H)\Pr(Y=y\mid C,H).
 \label{eq:terminal-policy-value}
\end{equation}
Thus $J_\mu=\mathbb E_{\mu,M_C}[R_\nu(H_B)\mid C,H_0]$, with $X$ also drawn from its conditional prior. Sampling $\hat Y\sim\nu$ and averaging $\eta(H_B,\hat Y)$ gives $V_0(H_B)-R_\nu(H_B)$. The subscript on $J_\mu$ suppresses $\nu$ only to simplify notation.

\paragraph{Exact enumeration and relabeling.}
For the five- and eight-bit tasks, the evaluator sums over all latent assignments and possible read returns. The channel may depend on history; the implemented tasks specify its independence and reliability explicitly. Legal regions are removed after reading. A bijection of region IDs preserves this recursion when applied to bindings, actions, histories and the channel together: the base posterior answer probability is unchanged, each action has the same corresponding outcomes, and induction on $b$ preserves $Q_b$ and $V_b$.

\paragraph{Cost of the exact reference.}
With $m$ legal regions and binary returns, a depth-$b$ search visits at most $\sum_{d=0}^{b}2^d(m)_d$ sequence nodes before memoization, where $(m)_d$ counts ordered distinct regions. Five regions and depth two give 91 nodes. Commuting histories can share states and reduce this count. This is evaluator computation, distinct from the observation budget, model calls and measured latency. The small controlled tasks permit exact enumeration; this count does not establish tractability for open-scene region proposals or learned read models.

%% file: agent_design_table.tex
\begin{table}[!htbp]
\centering\small
\setlength{\tabcolsep}{3pt}
\begin{tabular}{@{}p{.21\linewidth}p{.34\linewidth}p{.36\linewidth}@{}}
\toprule
Question & Evidence & Supported interpretation \\
\midrule
Conditional value & Five pinned open-weight VLMs; forty families and eight fixed parameter profiles & Reliability sensitivity and adaptation to a moving optimum are distinct. \\
Budget use & Forty high-gap paired anchors per configuration & No correct branch-to-selector transition under the tested direct-answer settings. \\
\rowcolor{paperBlue}
Complete observation policy & Three checkpoints; eight-region, three-read trajectories on forty families & Low first-action loss can coexist with later acquisition and terminal-use errors. \\
Natural visual evidence & Two eligible checkpoints on thirty real forms; six matched crop readers and whole-image controls & Crop selection and system accuracy differ; results depend on reference scoring. \\
Prospective terminal repair & Six fixed histories; development-locked diagnosis groups and two presentation controls & The selected repair predictions were not confirmed; absent diagnosis classes remain unidentifiable. \\
Presentation and reader controls & Historical paired inputs and shared-reader interventions & Exploratory effects locate stages of loss; they do not establish universal repair or isolated image causality. \\
\bottomrule
\end{tabular}
\caption{Questions, evidence and inferential limits. Families or documents are the sampling units; repeated decisions and shared-reader outputs are kept paired.}
\label{tab:diagnostic-design}
\end{table}

%% file: agent_protocol_summary.tex
\section{Protocol, Configuration and Reuse Summary}
\label{sec:protocol-summary}
Table~\ref{tab:protocol-summary} summarizes the hosted and earlier controlled protocols. GPT-4.1 and GPT-5 mini use the dated April 2025 GPT-4.1 and August 2025 GPT-5 mini snapshots. The enum wrapper requests a strict one-field JSON schema over the public legal menu. It applies no semantic repair; transport retries repeat the same request and increase actual request counts. The requested non-reasoning generation cap is 1,500 tokens at temperature 0. The GPT-5 route requests a 6,000-completion-token cap without a temperature parameter. Its effort is explicitly low in the boundary protocol and unspecified in the factorial protocol; ``default'' therefore denotes an endpoint setting, not a measured or matched amount of reasoning.

\begin{table*}[!htbp]
\centering\small
\setlength{\tabcolsep}{4pt}
\begin{tabular}{@{}p{.18\textwidth}p{.15\textwidth}p{.23\textwidth}rrp{.17\textwidth}@{}}
\toprule
Protocol / model & Reasoning / output & Split and reuse & \shortstack[r]{Scheduled\\units} & \shortstack[r]{Actual\\requests} & Outcome handling \\
\midrule
Factorial / GPT-4.1 & Not requested; strict enum & 6 dev, 48 new confirmation families; 9 states, 6 arms & 2,592 & 2,923 & 2,592 legal; no repairs or unresolved missing \\
Factorial / GPT-5 mini & Endpoint default; strict enum & Same frozen family/state/arm design & 2,592 & 2,592 & 2,591 legal, 1 invalid; no repairs or missing \\
Moving boundary / GPT-4.1 & Not requested; strict enum & 6 dev families, 4 dev profiles; 24 new families, 4 held-out profiles, 9 levels/profile & 864 & 880 & 864 legal; no repairs or unresolved missing \\
Moving boundary / GPT-5 mini & Low effort; strict enum & Same frozen held-out boundary design & 864 & 864 & 864 legal; no repairs or unresolved missing \\
Computed-value follow-up / GPT-4.1 & Not requested; strict enum & Reuses 96 boundary states at $\rho=.74$; two new aid arms & 192 & 192 & Baseline 96 states reused, not new independent tests \\
Downstream crop follow-up / GPT-4.1 & Not requested; strict enum & 24 reused masked families, 2 rules, paired arms; common first action/read & 48 episodes & 348 & 338 scored steps, all legal; 98 distinct crop reads \\
Initial Qwen legal menu & Instruct checkpoint; constrained choice & Separate 24-family holdout; two input modes & 1,056 & 1,056 & 1,056 legal first decoded actions \\
Initial Qwen free form & Instruct checkpoint; bounded repair & Same holdout; separate output protocol & 1,056 & 2,112 & 0 strict, 1,043 repaired, 13 invalid \\
\bottomrule
\end{tabular}
\caption{Confirmation accounting. A unit is a scheduled decision except in the multi-step crop row. Actual requests include transport attempts and, for free-form Qwen, the separate formatting repair. The 432-cell initial-decision scores elsewhere exclude forced histories; they are not the full 1,056-decision interface denominator. The factorial's 432 decisions per arm are nested in 48 families; the boundary's 864 decisions are nested in 24 families and four profiles.}
\label{tab:protocol-summary}
\end{table*}

The initial Qwen conditions use the same pinned revision, bfloat16 and a 256-token free-generation cap. Menu-constrained and repaired results remain distinct decoding conditions. The initial hosted map-ablation protocol permits one formatting repair and uses a different state schedule, introduction and output contract from the later enum protocols. Historical Sonnet and Qwen3.5 records are separate archives, not matched baselines for the new models.

Families are the resampling unit in the reported intervals, retaining all within-family states and arms. The four held-out boundary profiles were not used for predictive fitting, but the intervals condition on them. A constant family-bootstrap contrast can yield a degenerate interval; this describes the finite panel rather than establishing population certainty. The experiments did not prespecify an equivalence margin. Reused-state computational aids and same-family schema/ledger follow-ups are post-diagnostic, and cannot serve as independent confirmation of predicted repairs.

%% file: confirm_v2_appendix.tex
\section{Confirmation Stimulus and Interface Audit}
\label{sec:supp-v2}
The decision panel freezes a row, grid and scatter layout with independently assigned visual tags. The 22 actions per family comprise eight fair-prior cells crossing rule, budget and reassignment, two unequal-prior cells, eight reliability cells and four forced-history cells. At the two-reveal budget, selector reliability takes six levels from .5 to 1.0. The one-reveal comparison uses .5 and 1.0. We preserve role labels across the image-plus-map and map-only prompts. In the reliability prompts the selector window appears in the noise description. This information is public in all modalities. The response interface uses the same final-action delimiter, with one bounded formatting repair for a completed but unparseable reply. We report strict and repaired successes separately. Transport failures contribute missing observations. Table~\ref{tab:confirmation-v2} reports the held-out first-action results.

For the crop-return comparison, the original optically blurred overview retained a deterministic signal: a directional-gradient statistic recovered all 120 bit values in the 24 confirmation families. The replacement overviews render a gray patch independently of the hidden values while preserving the original crop images. Flipping all five bits leaves the overview pixels identical for each of the 27 development and confirmation families. The unchanged crop images retain the full directional code. The original overview version is omitted from perceptual comparisons.

\paragraph{Model and decoder settings.}
For GPT-4.1 the April 2025 snapshot was returned. The DeepSeek gateway returned its Flash system rather than the generic chat identifier used in the request. Both routes requested at most 1,500 output tokens. Qwen3-VL-2B-Instruct used a fixed revision recorded with the artifact, bfloat16, Transformers 5.21.0 and PyTorch 2.14.1. The legal-menu condition holds the task prompt fixed, prefills \texttt{FINAL:} and chooses from the valid windows. The separate free-form development check requested at most 256 tokens.

\begin{table}[!htbp]
\centering\footnotesize
\setlength{\tabcolsep}{3pt}
\begin{tabular}{@{}lrrrr@{}}
\toprule
Input & Exact hits & $S$ at .5 & $S$ at 1 & $F\to S$ \\
\midrule
\multicolumn{5}{@{}l}{\textbf{GPT-4.1}} \\
Image + map & 288/432 & 20/24 & 24/24 & 0/24 \\
Image only & 269/432 & 24/24 & 24/24 & 0/24 \\
Map only & 372/432 & 2/24 & 24/24 & 10/24 \\
\multicolumn{5}{@{}l}{\textbf{GPT-5 mini}} \\
Image + map & 409/432 & 0/24 & 24/24 & 18/24 \\
Map only & 401/432 & 1/24 & 24/24 & 19/24 \\
\multicolumn{5}{@{}l}{\textbf{DeepSeek-flash}} \\
Map only & 383/432 & 2/24 & 24/24 & 5/24 \\
\multicolumn{5}{@{}l}{\textbf{Qwen3-VL-2B, legal menu}} \\
Image + map & 205/432 & 21/24 & 21/24 & 0/24 \\
Map only & 206/432 & 18/24 & 20/24 & 0/24 \\
\multicolumn{5}{@{}l}{\textbf{Qwen3-VL-2B, repaired text}} \\
Image + map & 213/432 & 23/24 & 24/24 & 0/24 \\
Map only & 209/432 & 24/24 & 23/24 & 0/24 \\
\bottomrule
\end{tabular}
\caption{Held-out decisions on three layouts. Exact first hits exclude forced continuations. The mean uniform-action hit rate is 37.8\% for these 432 cells. The selector columns refer to the two-reveal reliability endpoints. The final column counts unequal-prior pairs whose one- and two-reveal actions follow the exact horizon prediction.}
\label{tab:confirmation-v2}
\end{table}

\paragraph{Masked crop replication and planner replacement.}
GPT-5 mini returned the August 2025 snapshot. Its 216 masked confirmation episodes completed with 456 strictly legal steps and no repairs. It read 15/24 masked-overview targets correctly (56.7\% balanced accuracy), matching the overall accuracy of the observed majority-bit baseline. It read 48/48 standalone crops correctly. Switch answer counts were 16/24 at budget one and 24/24 at budget two in both return modes. Two-reveal XOR yielded 23/24 with crops and 24/24 with text. All first actions were optimal in both switch budgets and in two-reveal XOR.

For the GPT-4.1 post-diagnostic one-reveal intervention, the planner used only the public map, rule and fair prior to select the first crop. It selected the first exact optimum in window order. The final-answer prompt format and model were unchanged and the requested crop was the only changed observation. All 24 trials completed in each arm. The native arm made 48 requests in total, while the planned arm made 24.

\paragraph{Document crop accounting.}
With two crops, direct GPT-4.1 and the decoupled controller each made 72 requests and reached 24/24 correct. Reported input-plus-output tokens were 126,776 and 61,312, respectively. Their cumulative request times were 306.61 and 224.38 seconds. The times sum individual request durations across concurrently scheduled tasks. In the matched routing audit, querying both price branches required 96 primary calls and 97 actual requests including one additional attempt. Each individual policy can be deployed with one grounding call and two crop reads.

\input{dense_figure}

\paragraph{Denser-layout design and scoring.}
Figure~\ref{fig:dense-examples} shows examples of the layouts. The 24 held-out families contain four instances of each layout-by-row-count pair. Eight families use each of status, amount and flag encodings. Six separate development families were used to check reply formatting. Each family has five candidate regions and repeats one target case ID across regions. In the full and overview probes the question identifies the same region and case as in the crop probe. We compare exact final field strings. Status output errors disagree with the correct status named in earlier response text; they remain errors under the common final-line scoring contract. The two-crop planner comparison uses the same crop decode when both policies request the same region and the same posterior answer rule. Table~\ref{tab:dense-results} reports the crop and controller comparisons. The independent OCR audit recovered all 120 region fields from native confirmation crops, including all 24 targeted crop fields.

\begin{table}[!htbp]
\centering\small
\begin{tabular}{@{}p{.43\columnwidth}rrr@{}}
\toprule
Dense scene condition & Correct & Requests & \shortstack[r]{Reported\\tokens}\\
\midrule
Exact field from overview & 19/24 & 24 & 17,899 \\
Exact field from native full page & 22/24 & 24 & 28,789 \\
Exact field from native crop & 22/24 & 24 & 12,265 \\
Answer from overview & 20/24 & 24 & 25,411 \\
Answer from native full page & 23/24 & 24 & 35,944 \\
One-step controller, two crops & 18/24 & 48 & 26,113 \\
Remaining-budget controller, two crops & 24/24 & 48 & 26,080 \\
Direct model, one crop & 15/24 & 48 & 60,511 \\
Direct model, one text return & 19/24 & 49 & 49,164 \\
Direct model, two crops & 24/24 & 72 & 102,367 \\
Direct model, two text returns & 24/24 & 72 & 70,579 \\
\bottomrule
\end{tabular}
\caption{Completed dense visual confirmation on the same 24 families across conditions. Request and token counts for each controller correspond to the reads used by that policy. When both controllers request a region, the audit can reuse its read for scoring.}
\label{tab:dense-results}
\end{table}

\paragraph{Structured status response after the diagnostic.}
We reran the eight status families after observing disagreements between the field named in the response and the final answer. Each request retained the view, region, case and field. It used a strict JSON enum for the status, with enum order counterbalanced before the rerun. The original and structured conditions yielded 3/8 versus 8/8 correct for overviews, 6/8 versus 8/8 for native full pages and 6/8 versus 8/8 for native crops. Across the 24 paired requests, nine changed from incorrect to correct and none changed from correct to incorrect. The paired gain was $+37.5$ percentage points (95\% interval $[12.5,66.7]$), resampling the eight families with their three views. These are same-family follow-up observations. The original exact-field results remain the primary comparison of views.

\paragraph{Free-form Qwen confirmation with repair.}
Under the separate free-form protocol, 0/1,056 confirmation responses passed strict first-pass parsing, and 1,043/1,056 produced a legal action after the single repair. The 13 remaining invalid responses count as misses. The endpoint shift in selector choices with the image and map was $+4.2$ pp ($[0,12.5]$), compared with $-4.2$ pp ($[-12.5,0]$) with the map alone. The rule-conditioned selector heuristic led the predictive comparison in both modes. This protocol mostly measures the actions extracted by a second request, while the legal-menu protocol constrains the first decoded answer.

\paragraph{GPT-5 mini decision confirmation.}
The 1,056 confirmation decisions produced 1,010 strict and 35 repaired legal actions, with 11 invalid actions. The three initial technical failures resolved on rerun. In both input modes the selector count shifted from 0/48 to 48/48 across fair switch budgets and stayed zero on XOR. At budget two, the image-plus-map reliability curve selected the selector in $0,12,20,24,24,24$ of 24 families in ascending reliability. The map-only curve selected it in $1,10,17,23,22,24$ of 24. The horizon reference had the lowest held-out log loss in both modes (.728 and .726 nats per action). Neither empirical curve places every action on the exact side of the crossover. At reliability .6, the selector is suboptimal even though it was selected on 12/24 image-plus-map and 10/24 map-only trials.

%% file: dense_figure.tex
\begin{figure}[!htbp]
\centering
\begin{minipage}[t]{.36\linewidth}\centering\small\textbf{Register}\\[3pt]
\includegraphics[width=\linewidth]{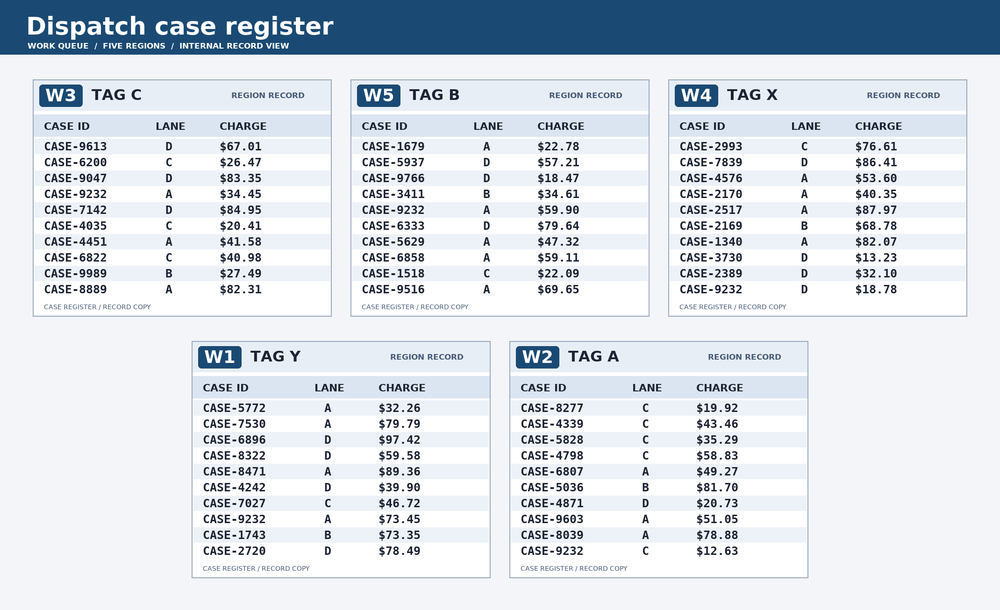}
\end{minipage}\hspace{2.5em}
\begin{minipage}[t]{.36\linewidth}\centering\small\textbf{Ledger}\\[3pt]
\includegraphics[width=\linewidth]{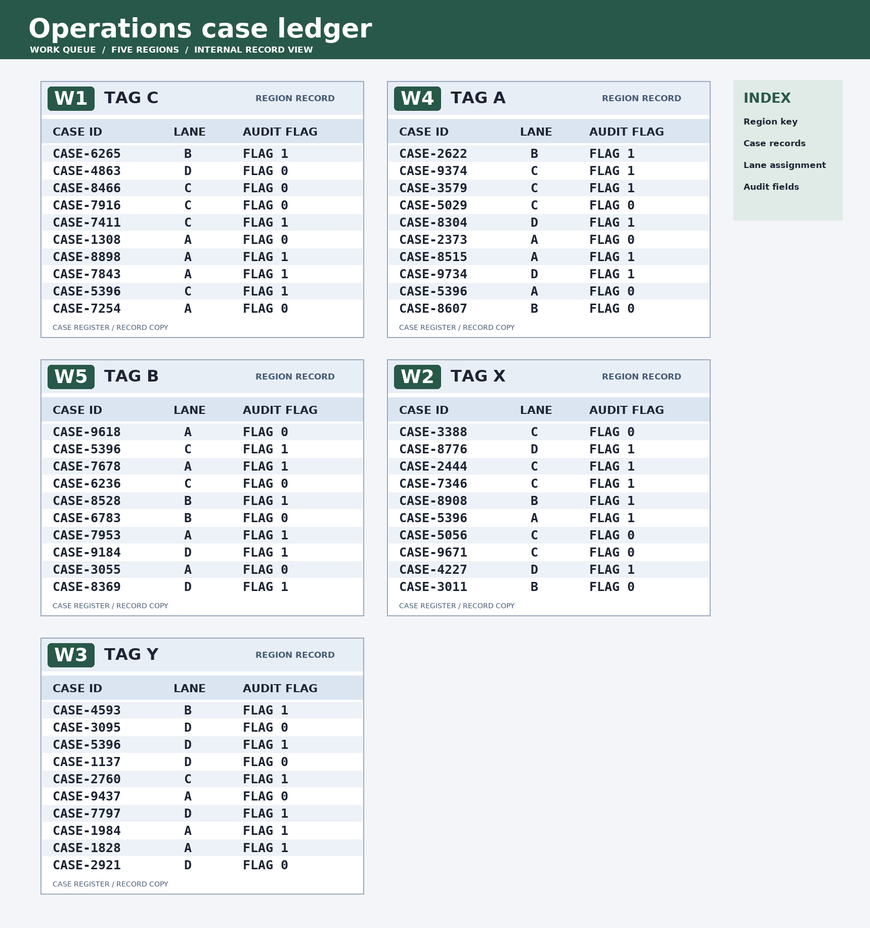}
\end{minipage}\\[6pt]
\begin{minipage}[t]{.36\linewidth}\centering\small\textbf{Dashboard}\\[3pt]
\includegraphics[width=\linewidth]{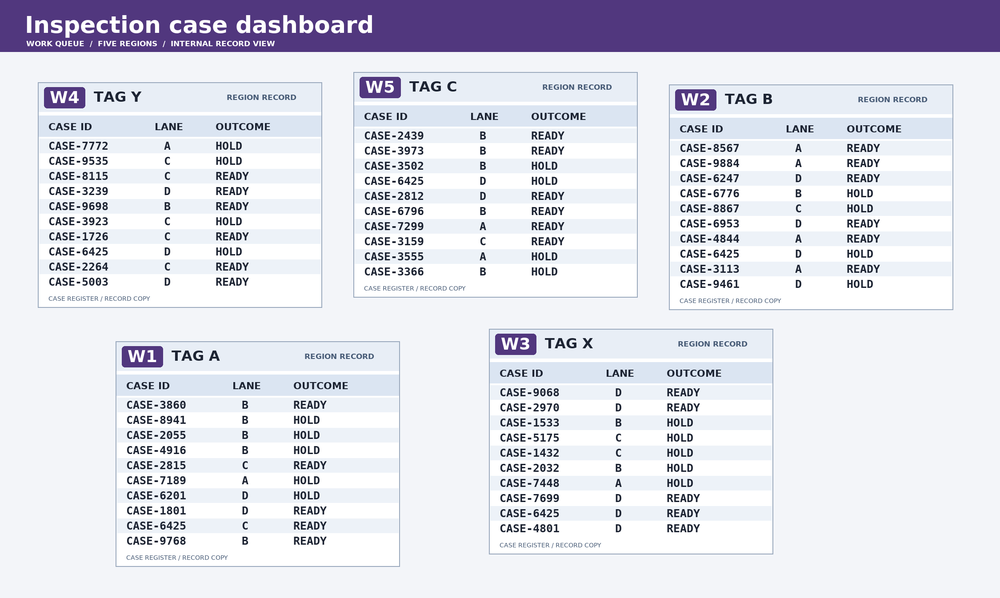}
\end{minipage}\hspace{2.5em}
\begin{minipage}[t]{.36\linewidth}\centering\small\textbf{Native ledger crop}\\[3pt]
\includegraphics[width=\linewidth]{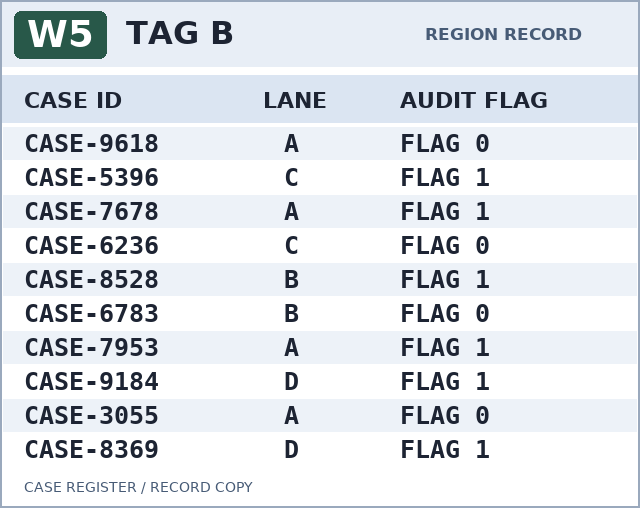}
\end{minipage}
\caption{The first ten-row development example from each layout and an unmodified native-resolution crop from the illustrated ledger. Overviews downsample the complete images. Repeated case labels occur among distractor rows; the requested field specifies the latent bit.}
\label{fig:dense-examples}
\end{figure}

%% file: agent_factorial_appendix.tex
Table~\ref{tab:factorial-gpt4} reports the GPT-4.1 contrasts. The experiment freezes 48 confirmation families across three rendered layouts after a six-family development check. Each family has nine matched states: fair switch at one and two reveals, unequal priors at one and two, fair XOR at two, low and high selector reliability at two, and two forced-history states with opposite selector bits. Region row order and legal menu order are randomized within family, and field order is balanced. The two presentations share the same rule and action schema. The task image is identical in the image arms and contains no hidden bit values. The auxiliary neutral image matches its dimensions and contains no region information. The appearance control places facts in a field-indexed table, independently ordering window IDs within each field.

For example, in one development scene, W1 carries tag B and has the selector role. Its prior is .50 and its read reliability is 1.00. It has not been opened, so its returned bit is unknown and its status is \texttt{not returned}. The ledger puts these facts in a single W1 row. Prose organizes the same entries under separate field names. Neither condition supplies derived quantities. An independent parser verifies the public state and reference values from the rendered input.

Each model is evaluated with a strict one-field JSON enum over that state's legal actions. A transport retry repeats the same request. A completed invalid reply ends the episode. The primary loss contrast uses common legal pairs and 2,000 family bootstrap resamples. In a sensitivity analysis an invalid action gets zero expected reward, or the value of the worst legal action. Optimal-action hits treat invalid replies as misses. Technical failures and request costs are counted separately.

\begin{table}[!htbp]
\centering\small
\begin{tabular}{@{}p{.43\columnwidth}rrr@{}}
\toprule
GPT-4.1 presentation & \shortstack[r]{Legal /\\scheduled} & \shortstack[r]{Exact\\hits} & \shortstack[r]{Mean\\loss} \\
\midrule
Prose, no image & 432/432 & 274/432 & .0542 \\
Ledger, no image & 432/432 & 276/432 & .0519 \\
Prose, task image & 432/432 & 280/432 & .0473 \\
Ledger, task image & 432/432 & 284/432 & .0426 \\
Prose, neutral image & 432/432 & 279/432 & .0484 \\
Field-indexed table, task image & 432/432 & 278/432 & .0496 \\
\bottomrule
\end{tabular}
\caption{Factorial and auxiliary GPT-4.1 comparisons. Mean loss uses legal choices. No confirmation action remained technically missing after reruns.}
\label{tab:factorial-gpt4}
\end{table}

With the same prose, adding the task image rather than a neutral image changed loss by $-0.12$ pp ($[-0.58,0.46]$). With the task image fixed, the region ledger versus the field-indexed table changed loss by $-0.69$ pp ($[-1.39,0.00]$). The per-region ledger preserves the relation between a region and all of its fields. These contrasts are smaller than the old map ablation, which used a different protocol. The image-plus-ledger arm used 442,800 reported tokens, versus 485,136 for image-plus-prose, over the 432 completed decisions in each arm.

\IfFileExists{agent_factorial_appendix_gpt5.tex}{\input{agent_factorial_appendix_gpt5}}{}

%% file: agent_factorial_appendix_gpt5.tex
\begin{table}[!htbp]
\centering\small
\begin{tabular}{@{}p{.43\columnwidth}rrr@{}}
\toprule
GPT-5 mini presentation & \shortstack[r]{Legal /\\scheduled} & \shortstack[r]{Exact\\hits} & \shortstack[r]{Mean\\loss} \\
\midrule
Prose, no image & 432/432 & 421/432 & .0016 \\
Ledger, no image & 431/432 & 422/432 & .0009 \\
Prose, task image & 432/432 & 419/432 & .0029 \\
Ledger, task image & 432/432 & 414/432 & .0025 \\
Prose, neutral image & 432/432 & 421/432 & .0011 \\
Field-indexed table, task image & 432/432 & 415/432 & .0022 \\
\bottomrule
\end{tabular}
\caption{Factorial and auxiliary GPT-5 mini comparisons under the endpoint's default reasoning setting. Exact hits include the one invalid completion as a miss. Mean loss uses legal choices.}
\label{tab:factorial-gpt5}
\end{table}

In Table~\ref{tab:factorial-gpt5}, the invalid reply was in the ledger-only arm. Scoring it as zero expected reward changes the paired no-image ledger effect from $-0.07$ to $+0.11$ pp ($[-0.19,0.54]$). With the same prose, the task versus neutral image changes loss by $+0.17$ pp ($[-0.05,0.47]$). With the task image fixed, the ledger versus field-indexed table changes loss by $+0.03$ pp ($[-0.16,0.24]$). The fair-switch budget response with no image rose by 97.9 pp in prose and 100 pp in the ledger, a $+2.08$ pp paired change ($[0,6.25]$). Across both models, the same-image ledger contrasts in action loss and fair-switch budget response include zero.

%% file: review3_distribution_table.tex
\begin{table*}[!htbp]
\centering\small
\setlength{\tabcolsep}{4pt}
\begin{tabular}{@{}llrrrrrrrr@{}}
\toprule
Model & Input & Legal/planned & Hits & Mean & Median & IQR & P90 & $R\leq .01$ & $R\leq .05$ \\
\midrule
GPT-4.1 & Prose + image & 432/432 & 280 & 4.73 & 0.00 & 4.40 & 25.00 & 64.8 & 87.0 \\
GPT-4.1 & Ledger + image & 432/432 & 284 & 4.26 & 0.00 & 4.40 & 25.00 & 65.7 & 88.0 \\
GPT-4.1 & Prose, no image & 432/432 & 274 & 5.42 & 0.00 & 4.82 & 25.00 & 63.4 & 85.6 \\
GPT-4.1 & Ledger, no image & 432/432 & 276 & 5.19 & 0.00 & 4.50 & 25.00 & 63.9 & 86.1 \\
\addlinespace
GPT-5 mini & Prose + image & 432/432 & 419 & 0.29 & 0.00 & 0.00 & 0.00 & 97.0 & 99.5 \\
GPT-5 mini & Ledger + image & 432/432 & 414 & 0.25 & 0.00 & 0.00 & 0.00 & 95.8 & 99.5 \\
GPT-5 mini & Prose, no image & 432/432 & 421 & 0.16 & 0.00 & 0.00 & 0.00 & 97.5 & 99.8 \\
GPT-5 mini & Ledger, no image & 431/432 & 422 & 0.09 & 0.00 & 0.00 & 0.00 & 97.9 & 100.0 \\
\addlinespace
\bottomrule
\end{tabular}
\caption{Retrospective loss distributions over the nine fixed states in each of 48 families. Mean, median, IQR and P90 are action regret in percentage points; the final two columns are percentages of legal actions. Hit counts use all planned decisions, with invalid replies as misses; numeric loss and cumulative proportions use legal replies. The GPT-5 mini no-image ledger arm has one invalid reply and no technical missing trial. There are 288 unique-optimum and 144 tied-optimum decisions in each planned arm; the mean uniform-menu hit baseline is 27.8\%. Quantiles use linear interpolation at $(n-1)p$. These empirical distributions do not establish equivalence or uncertainty over a new state distribution.}
\label{tab:review3-distributions}
\end{table*}

%% file: agent_boundary_appendix.tex
\begin{table}[!htbp]
\centering\small
\setlength{\tabcolsep}{2pt}
\begin{tabular}{@{}p{.27\columnwidth}rrrrr@{}}
\toprule
Model & \shortstack[r]{Exact\\hits} & \shortstack[r]{Mean loss\\(accuracy)} & \shortstack[r]{$\rho$\\only} & \shortstack[r]{$D$\\only} & $\rho$, $p$, rB \\
\midrule
GPT-4.1 & 480/864 & .00777 & .00484 & .00499 & .00487 \\
GPT-5 mini, low effort & 567/864 & .00490 & .21044 & .25441 & .20781 \\
\bottomrule
\end{tabular}
\caption{Moving-boundary confirmation and predictive loss. The last three columns are held-out log loss in nats per binary selector choice. All 864 action replies were legal for each model.}
\label{tab:boundary-confirmation}
\end{table}

At the common $\rho=.74$, GPT-5 mini chose the selector in 23, 23, 24 and 22 families out of 24 across the profiles in Table~\ref{tab:boundary-profiles}. The discrete action fits favored the selector-role policy over the exact horizon model for both GPT-4.1 (.008 versus 1.305 nats/action) and GPT-5 mini (.509 versus 1.184). This comparison fits only the lapse rate on development data. The main figure displays the same choices with the four value crossings aligned to zero.

%% file: agent_presentation.tex
\paragraph{The same binding supports different policies.}
On 24 held-out families, GPT-4.1's map-only input raised first-action hits from 288/432 to 372/432 ($+19.4$ pp, 95\% family interval $[17.6,21.3]$), reducing mean loss from .0340 to.0053. With image plus map it chose the selector in all 48 fair switch trials at either budget; map-only counts were 4/48 and 48/48. A separate binding probe recovered 24/24 scenes. This contrast also changes the scene introduction. GPT-5 mini's difference was$-1.9$ pp ($[-4.2,.5]$); full conditions and historical checkpoint results are in the supplement.

\IfFileExists{agent_factorial.tex}{\input{agent_factorial}}{}

%% file: agent_factorial.tex
\input{agent_factorial_methods}

\paragraph{Reorganization leaves a fixed choice at the root.}
For GPT-4.1, the region ledger with the same image changed mean action loss from .0473 to .0426 ($-0.46$ pp, 95\% family interval $[-1.27,0.23]$, 432 legal pairs). Without an image, the corresponding difference was $-0.23$ pp ($[-0.93,0.46]$). The interaction was $-0.23$ pp ($[-1.16,0.58]$). The seven initial-state contrasts had zero mean change in action loss. In particular, GPT-4.1 selected $S$ on every fair switch trial at both budgets and at both reliability endpoints in every arm. Any observed mean change came from the two forced-history states. The neutral-image and table-appearance comparisons provide no clear evidence that visual content or the appearance of a table alone accounts for the result.

The earlier gain from the map-only input did not recur under this protocol, which changes the state panel, wording and output contract together. The factorial isolates organization of the same facts under one interface. Whether it helps the agent use evidence in a full tool loop is a separate test (the main-paper intervention section).

The two models have different headroom. GPT-4.1's prose-with-image loss is .0473, whereas GPT-5 mini's is .0029, already near the exact reference. The former leaves acquisition loss that reorganization did not clearly remove; the latter constrains the possible gain. No equivalence margin was prespecified.

\IfFileExists{agent_factorial_gpt5.tex}{\input{agent_factorial_gpt5}}{}
\IfFileExists{review3_state_figure.tex}{\input{review3_state_figure}}{}

%% file: agent_factorial_methods.tex
\paragraph{Factorial test of public state organization.}
A new paired panel crosses image presence with organization of the same public facts. We freeze 48 confirmation families after checking six separate development families. Each confirmation family supplies nine states that distinguish budget response, asymmetric priors, XOR, stated read reliability and forced continuations. The prose groups facts by field, while the ledger groups the same facts by region. Both give each region's tag, role, observation status, returned bit, validity, prior and reliability. The rule, budget, legal menu and strict JSON action contract are shared. Unobserved values remain unknown, and neither presentation gives a posterior, action value or recommendation. An automated round-trip check recovers the same task and exact action values from every development and confirmation rendering.

The primary contrast is ledger minus prose action loss on common legal choices with the \emph{same} attached image. We also compare the two presentations without the image and estimate the interaction. The same loss scale compares states with tied optima without penalizing a switch between equally useful actions. We report optimal-action hits with invalid replies as misses and track transport failures separately. All intervals resample families with their conditions intact.

%% file: agent_factorial_gpt5.tex
\paragraph{GPT-5 mini adapts to the budget in both presentations.}
With the same image, the ledger changed action loss by $-0.04$ pp ($[-0.36,0.23]$, 432 legal pairs). Without the image, the change was $-0.07$ pp ($[-0.22,0.06]$, 431 legal pairs). The interaction was $+0.04$ pp ($[-0.30,0.37]$). The fair-switch selector rate rose by 97.9 pp between one and two reveals in both image arms, giving a ledger effect of 0 pp ($[-6.25,6.25]$). The corresponding reliability endpoint contrast rose by 97.9 pp in each image arm. With the image, exact hits fell from 419/432 to 414/432 while mean loss decreased from .0029 to .0025: hit rate measures optimality, whereas loss also weights the severity of a miss. Absolute losses and paired differences therefore need to be read together (Fig.~\ref{fig:agent-state}).

%% file: review3_state_figure.tex
\begin{figure}[!htbp]
\centering
\includegraphics[width=.62\linewidth]{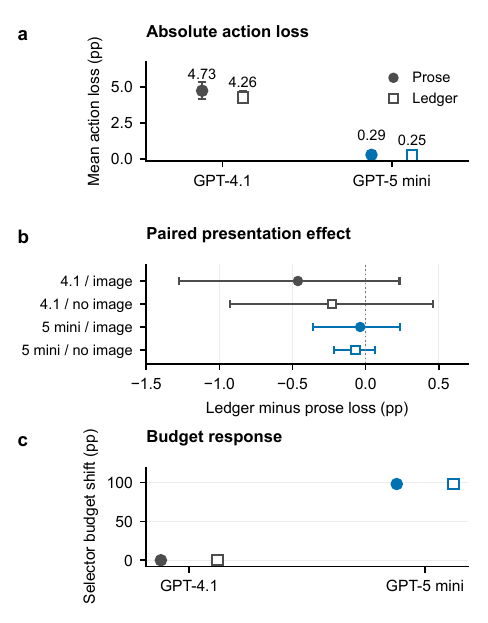}
\caption{Absolute performance changes the interpretation of a small presentation effect. \textbf{a}, Mean action loss with the same task image: GPT-4.1 has room to improve, whereas GPT-5 mini starts near the optimum. \textbf{b}, Paired ledger-minus-prose loss, with and without the image; circles and squares indicate image and no-image conditions. \textbf{c}, Fair-switch selector-rate change from one to two reveals, with the image held fixed. In a and c, filled circles denote prose and open squares the ledger. Model colors are consistent across panels. Intervals are 95\% family bootstraps over 48 families, preserving all nine states per family. Numeric loss uses legal actions; the no-image GPT-5 mini contrast has 431/432 legal pairs. Intervals containing zero do not establish equivalence.}
\label{fig:agent-state}
\end{figure}

%% file: legacy_intervention_details.tex
\paragraph{Prospective repair was not confirmed.}
The terminal test fixes all six histories before confirmation and compares prose, ledger and an active-field control. Diagnosis groups are locked by a development rule. Qwen3-VL-8B has zero overall prose--ledger gain and specificity of -3.00 pp (98.75\% interval [-20.51,13.49]); its development-selected repair prediction was not confirmed. InternVL, MiniCPM fast and Qwen3-VL-32B have no development-positive state, so specificity is unidentifiable. Intervals resample forty acquisition families or 39 terminal components after duplicate-input merging; their scope is the fixed parameter/state panel.

\paragraph{Match the reader before comparing policies.}
On 24 denser generated scenes, the remaining-budget controller answered 24/24 versus 18/24 for a one-step controller sharing its crop reader, decoder and two-reveal budget (+25.0 pp, 95\% interval [8.3,41.7]). Direct GPT-4.1 also reached 24/24; native whole-image reading reached 23/24. Thus the controller contrast is consistent with a lookahead advantage under the matched-reader comparison, while the competitive direct systems constrain its practical advantage. A separate routing audit shared the binding, selector and price readers: adaptive routing achieved 24/24 versus 12/24 in expectation for uniform fixed-branch choice. Counterfactual audit reads are separate from deployment calls.

\paragraph{A post-diagnostic format intervention.}
A same-sample, post-diagnostic status enum raised exact outputs from 15/24 to 24/24 across three views on eight status families (+37.5 pp, [12.5,66.7]). It targets delivery of the observed value. The independent open-checkpoint test provides the prospective repair check and retains its unconfirmed predictions. Full historical crop, controller, token and output-contract comparisons are in the supplement.

%% file: open_model_methods.tex
\section{New Matched-Protocol Open-Model Confirmation}
\label{sec:new-open-models}
This panel is separate from the historical model archives. It evaluates conditional decisions from a supplied region map and public task model, without image input. It therefore measures the policy component of the multimodal checkpoints under a controlled text interface.

Each configuration has 180 development acquisition queries and 108 development terminal queries, followed by 2,320 and 720 confirmation queries, respectively: at most 3,328 requests. Development completion and at least 95\% strict legality are required before confirmation. This is an interface gate, not a policy-accuracy gate. Each trial permits one attempt; technical failures consume budget, remain in the scheduled denominator, and are distinguished from invalid responses. No semantic repair is applied.

The acquisition panel crosses forty held-out binding families with eight parameter profiles and seven reliability values per profile (2,240 queries). Eighty additional fair high-gap anchors compare one- and two-observation budgets. Anchor parameters also occur in development. Non-anchor mean loss and high-gap anchor loss are reported separately. The adaptation contrast holds selector reliability at .74 and compares selector rates across profiles with positive and negative selector advantage. For the exact remaining-budget reference, this contrast is 100 percentage points. Family intervals condition on the eight fixed profiles. Six-decimal public probabilities were checked against full-precision values: neither action rankings nor values changed on the frozen query set.

The terminal panel fixes six public evidence states and three presentations (prose, ledger and active-field control) per family. The development rule selects diagnosis-positive states only when prose loss exceeds .01 and ledger loss is lower than both alternatives. The rule and each model's resulting state groups are locked before terminal confirmation. All states remain in the analysis, including models with no positive or no negative group; specificity is unidentifiable in those cases. One public tie-state triple is duplicated across families. Confirmation deduplicates it, giving 239 unique triples and 39 connected family components; the unmerged family analysis is a sensitivity check.

The four predeclared contrasts per model are acquisition adaptation, prose--ledger gain, active-control--ledger gain and diagnosis specificity. The last contrast is the gain difference between the development-positive and development-negative groups. Each uses 20,000 paired family/component bootstrap draws with the fixed analysis seed and a 98.75\% percentile interval; descriptive acquisition loss also has a 95\% interval. Intervals and non-estimable entries are retained without selecting models or states by confirmation effects. Between-model comparisons are descriptive; model-specific diagnosis groups can differ.

The current direct-answer conditions use pinned checkpoints, BF16 weights without quantization, the official chat interfaces, temperature zero and top-$p$ one. The requested output cap is 4,096 tokens. Strict JSON constrains generation from the first token to the legal final-answer candidates, making the effective cap much shorter (seven generated tokens for the completed Qwen3-VL-8B configuration). Strict legality is partly enforced by decoding. Full source/configuration hashes, checkpoint revisions, chat-template hashes, hardware, preprocessing, raw responses, stop reasons, measured tokens and request durations accompany each configuration. Summed request durations are reported separately from end-to-end latency.

\subsection{Configurations and Execution Coverage}
Table~\ref{tab:model-components} lists the checkpoints and their vision and language components. Tables~\ref{tab:new-open-coverage} and~\ref{tab:new-open-cost} give scheduled and completed requests, measured tokens and request durations for every configuration.

\input{model_components}
\input{open_model_coverage}
\input{open_cost_details}
\subsection{Acquisition and Paired Budgets}
Table~\ref{tab:new-open-models} reports acquisition loss and adaptation, Table~\ref{tab:open-budget-pairs} the paired anchor transitions, and Table~\ref{tab:anchor-loss-intervals} the anchor-loss intervals.

\input{open_acquisition_details}
\input{open_budget_pairs}
\input{open_anchor_intervals}
\subsection{Terminal Presentation and Diagnosis Groups}
Tables~\ref{tab:new-open-terminal} and~\ref{tab:terminal-group-gains} give the terminal contrasts overall and within the development-selected diagnosis groups.
\input{open_terminal_details}
\input{terminal_group_gains}

\paragraph{A prospective negative result.}
Qwen3-VL-8B-Instruct completed all 3,328 planned requests with strict legal outputs and no transport failures or recorded truncations. Its non-anchor mean acquisition loss was .6402 pp, but the fair one-observation anchors had zero optimal choices out of forty and 25 pp mean loss. The two-observation anchors had 39/40 optimal choices and .625 pp mean loss. A small non-anchor average therefore coexists with costly anchor errors. The prospective prose--ledger gain was zero, and the specificity interval included both positive and negative effects (Table~\ref{tab:new-open-terminal}). The development-selected repair prediction is therefore not confirmed; the positive active-control--ledger contrast is a separate endpoint.

InternVL3.5-8B and MiniCPM-V4.5 fast also completed all planned requests. Neither has a development-positive diagnosis state under the fixed selection rule, so their specificity entries are unidentifiable. Their overall prose--ledger point estimates are both $-2.92$ pp, with 98.75\% intervals that cross zero. No completed direct-answer configuration shows a broad terminal repair.

\paragraph{The larger Qwen3-VL checkpoint.}
The 32B condition completed all 3,328 requests without technical failures or truncation. It chose the selector on every acquisition-confirmation query, with .5963 pp non-anchor loss, zero fixed-reliability adaptation and 12.50 pp anchor loss. The acquisition family intervals are degenerate because these role-aligned choices are identical across the sampled families. Its prose--ledger gain is $-1.67$ pp ($98.75\%$ interval $[-4.27,.43]$), and active-control--ledger gain is $-2.08$ pp ($[-4.58,-.40]$). All development diagnosis states are negative, leaving specificity unidentifiable. The comparison extends checkpoint scale coverage under the same direct-answer protocol.

\subsection{Program References and the Scale of Loss}
Table~\ref{tab:core-program-references} scores fixed and one-step program choices on all frozen acquisition inputs, using the full tied action set with uniform weights. These exact task references were added after initial model collection and are not fitted to model responses. Non-anchor loss keeps the original equal-cell weighting, and adaptation keeps the fixed-reliability contrast. A constant selector has .5963 pp non-anchor loss, 12.50 pp anchor loss and zero adaptation. It therefore attains a low near-boundary average without tracking the moving optimal action. Fixed T denotes the branch active when the selector is one. Uniform random and tied-criterion values are exact averages rather than a seeded random realization. These are first-action opportunities under optimal continuation, not complete program-policy rollouts or matched-compute learned baselines.

\input{core_program_references}

\paragraph{Paired anchor choices.}
The anchor pairs preserve each family's bindings and menu while changing only the observation budget. The correct transition is a branch at budget one to the selector at budget two. Qwen3-VL-4B, Qwen3-VL-8B, Qwen3-VL-32B, InternVL3.5-8B and MiniCPM-V4.5 fast each produce zero such transitions among forty pairs. Selector-to-selector choices occur in 24, 39, 40, 40 and 35 pairs, respectively; all other role transitions remain in the released per-family table. Table~\ref{tab:open-budget-pairs} gives the complete paired categories; absolute losses appear in the main results table. These descriptive counts prevent a change in marginal optimal-action rates across budgets from being mistaken for the required within-family strategy change.

\subsection{Bounded Explicit-Thinking Condition}
The MiniCPM deep backend uses the same checkpoint and official thinking-mode switch, greedy beam-one generation and repetition penalty one. Before a naturally generated closing-think token, the vocabulary is unrestricted; only the final JSON is constrained. The 4,096-token cap covers reasoning, the boundary, final answer and EOS, with no forced closure or second generation. The transport timeout is 600 seconds, versus 180 for fast, to permit generation within the fixed token limit.

The deep condition stopped after eleven acquisition-development requests: ten reached the token limit without a final answer and one naturally completed. Ten nonlegal outputs make the original 95\% gate mathematically unreachable: even if all 169 remaining replies were legal, the total could only be 170/180 rather than the required 171. Terminal development and all confirmation queries were therefore unrequested. The feasibility-stop plan was recorded during development; it changes neither the fixed threshold nor the eventual eligibility verdict. This partial condition contributes no confirmation effect or interval.

It generated 41,760 tokens (41,752 in the reasoning segment) with 2,660.13 summed request seconds. Independent CPU reconstruction from the saved token IDs verified all eleven raw streams, template prefixes, boundaries and usage records. The successful reply also verifies that final-only constrained generation is executable. The official sampling defaults differ from the tested matched-greedy condition. The mode-comparison plan was recorded after fast data collection and initial deep development, before any deep confirmation, and is explicitly marked ineligible because deep never passed its gate.

%% file: model_components.tex
\begin{table}[!htbp]
\centering\small
\setlength{\tabcolsep}{3pt}
\begin{tabular}{@{}p{.29\linewidth}p{.64\linewidth}@{}}
\toprule
Configuration / component & Details \\
\midrule
\multicolumn{2}{@{}l}{\textbf{Qwen3-VL-4B}} \\
Visual / language & Qwen3-VL ViT / Qwen-derived text stack \\
Class / mode & 4B / Instruct, final JSON \\
\multicolumn{2}{@{}l}{\textbf{Qwen3-VL-8B}} \\
Visual / language & Qwen3-VL ViT / Qwen-derived text stack \\
Class / mode & 8B / Instruct, final JSON \\
\multicolumn{2}{@{}l}{\textbf{Qwen3-VL-32B}} \\
Visual / language & Qwen3-VL ViT / Qwen-derived text stack \\
Class / mode & 32B / Instruct, final JSON \\
\multicolumn{2}{@{}l}{\textbf{InternVL3.5-8B}} \\
Visual / language & InternViT / Qwen3-8B \\
Class / mode & 8B / Default, final JSON \\
\multicolumn{2}{@{}l}{\textbf{MiniCPM-V4.5}} \\
Visual / language & SigLIP / Qwen3-derived \\
Class / mode & 8B class / Fast, final JSON \\
\bottomrule
\end{tabular}
\caption{Five pinned configurations and their component families. Nominal classes follow checkpoint names. Core inputs are verified textual states. Version, decoding and hardware records accompany the appendix protocols.}
\label{tab:model-components}
\end{table}

%% file: open_model_coverage.tex
\begin{table}[!htbp]
\centering\footnotesize
\setlength{\tabcolsep}{3pt}
\begin{tabular}{@{}lrr@{}}
\toprule
Phase / status & Select & Answer / outcome \\
\midrule
\multicolumn{3}{@{}l}{\textbf{Qwen3-VL-4B / instruct}} \\
Development & 180/180 & 108/108 \\
Confirmation & 2320/2320 & 720/720 \\
Invalid / errors & 0 / 0 & completed \\
\multicolumn{3}{@{}l}{\textbf{Qwen3-VL-8B / instruct}} \\
Development & 180/180 & 108/108 \\
Confirmation & 2320/2320 & 720/720 \\
Invalid / errors & 0 / 0 & completed \\
\multicolumn{3}{@{}l}{\textbf{Qwen3-VL-32B / instruct}} \\
Development & 180/180 & 108/108 \\
Confirmation & 2320/2320 & 720/720 \\
Invalid / errors & 0 / 0 & completed \\
\multicolumn{3}{@{}l}{\textbf{InternVL3.5-8B / default}} \\
Development & 180/180 & 108/108 \\
Confirmation & 2320/2320 & 720/720 \\
Invalid / errors & 0 / 0 & completed \\
\multicolumn{3}{@{}l}{\textbf{MiniCPM-V4.5 / fast}} \\
Development & 180/180 & 108/108 \\
Confirmation & 2320/2320 & 720/720 \\
Invalid / errors & 0 / 0 & completed \\
\multicolumn{3}{@{}l}{\textbf{MiniCPM-V4.5 / deep}} \\
Development & 1/180 & 0/108 \\
Confirmation & 0/2320 & 0/720 \\
Invalid / errors & 10 / 0 & gate failed \\
\bottomrule
\end{tabular}
\caption{Coverage of locally saved core configurations, including incomplete and failed conditions. All use strict JSON unless an alternative interface is labeled. Select and Answer denote acquisition and terminal decisions. Development columns count legal outputs over planned queries; confirmation columns count attempts over planned queries, irrespective of legality. Invalid and transport/contract error counts pool all four panels. Unrequested trials are not technical failures. A development gate is unattainable once even all remaining legal outputs cannot reach 95\%; such a condition has no confirmation effect estimate. Only complete configurations enter the preceding confirmation and cost tables. The acquisition/terminal development caps are 180/108 and confirmation caps 2,320/720 per condition.}
\label{tab:new-open-coverage}
\end{table}

%% file: open_cost_details.tex
\begin{table}[!htbp]
\centering\footnotesize
\setlength{\tabcolsep}{3pt}
\begin{tabular}{@{}lrrr@{}}
\toprule
GPUs / calls & Input tokens & Output tokens & \shortstack{Seconds /\\truncated} \\
\midrule
\multicolumn{4}{@{}l}{\textbf{Qwen3-VL-4B / instruct}} \\
1 / 3328 & 871,120 & 22,468 & 1,010 / 0 \\
\multicolumn{4}{@{}l}{\textbf{Qwen3-VL-8B / instruct}} \\
1 / 3328 & 871,120 & 22,468 & 1,226 / 0 \\
\multicolumn{4}{@{}l}{\textbf{Qwen3-VL-32B / instruct}} \\
4 / 3328 & 871,120 & 22,468 & 3,517 / 0 \\
\multicolumn{4}{@{}l}{\textbf{InternVL3.5-8B / default}} \\
1 / 3328 & 994,256 & 22,468 & 1,470 / 0 \\
\multicolumn{4}{@{}l}{\textbf{MiniCPM-V4.5 / fast}} \\
1 / 3328 & 884,432 & 22,468 & 1,174 / 0 \\
\bottomrule
\end{tabular}
\caption{Measured cost of the complete map-only development and confirmation configurations (3,328 requests each). Hardware is NVIDIA RTX 3090 with 24 GiB per GPU; weights are BF16 without quantization. GPU counts follow the actual model device map, with no CPU/disk offloading. All 3,328 saved usage records per configuration are measured: 2,500 actions emit seven tokens and 828 terminal replies emit six, yielding 22,468 in every row. The fixed JSON trees explain these identical totals. Token and duration sums require measurement on every attempted request; otherwise the cell is omitted. Request seconds sum the client-measured request durations, including local transport, and are neither GPU kernel time nor concurrent end-to-end latency. The requested cap is 4,096 output tokens per call; effective limits and actual reasoning behavior differ by deployment condition and are recorded in its receipt. These figures exclude weight download, model loading and offline analysis.}
\label{tab:new-open-cost}
\end{table}

%% file: open_acquisition_details.tex
\begin{table}[!htbp]
\centering\footnotesize
\setlength{\tabcolsep}{3pt}
\begin{tabular}{@{}lrrr@{}}
\toprule
\shortstack{Legal / optimal\\(of 2,320)} & Anchor & \shortstack{Non-anchor\\loss (pp)} & \shortstack{Adaptation\\(pp)} \\
\midrule
\multicolumn{4}{@{}l}{\textbf{Qwen3-VL-4B / instruct}} \\
2320 / 1153 & 15.62 & \shortstack{0.85\\{}[0.69, 1.03]} & \shortstack{-0.62\\{}[-5.00, 3.12]} \\
\multicolumn{4}{@{}l}{\textbf{Qwen3-VL-8B / instruct}} \\
2320 / 1160 & 12.81 & \shortstack{0.64\\{}[0.58, 0.74]} & \shortstack{-0.62\\{}[-2.50, 0.00]} \\
\multicolumn{4}{@{}l}{\textbf{Qwen3-VL-32B / instruct}} \\
2320 / 1160 & 12.50 & \shortstack{0.60\\{}[0.60, 0.60]}$\dagger$ & \shortstack{0.00\\{}[0.00, 0.00]}$\dagger$ \\
\multicolumn{4}{@{}l}{\textbf{InternVL3.5-8B / default}} \\
2320 / 1171 & 12.50 & \shortstack{0.59\\{}[0.58, 0.60]} & \shortstack{0.00\\{}[0.00, 0.00]}$\dagger$ \\
\multicolumn{4}{@{}l}{\textbf{MiniCPM-V4.5 / fast}} \\
2320 / 1157 & 13.75 & \shortstack{0.62\\{}[0.59, 0.67]} & \shortstack{0.00\\{}[0.00, 0.00]}$\dagger$ \\
\bottomrule
\end{tabular}
\caption{New matched-protocol acquisition confirmation. Models use the same forty binding families, eight held-out parameter profiles and strict JSON interface. Optimal credits every tied optimum among all 2,320 queries; uniform-menu agreement is 30\%. Anchor loss is the descriptive mean over both budget conditions of the 80 fair high-gap anchor queries; the anchor parameters are shared with development. Non-anchor loss averages the 56 other cells with a 95\% family interval. Adaptation is the selector-rate difference between positive- and negative-value profiles at fixed reliability .74 (exact reference: 100 pp; predeclared 98.75\% interval). Effects and losses are in percentage points. $\dagger$ denotes a finite-panel degenerate interval. Direct constrained answers are not explicit-thinking tests; strict legality is partly enforced by decoding. Historical hosted runs use other protocols.}
\label{tab:new-open-models}
\end{table}

%% file: open_budget_pairs.tex
\begin{table}[!htbp]
\centering
\small
\begin{tabular}{lrrr}
\toprule
Model & $S\to S$ & $T/F\to S$ & Other pairs \\
\midrule
Qwen3-VL-4B & 24/40 & \cellcolor{paperAmber}0/40 & 16/40 \\
Qwen3-VL-8B & 39/40 & \cellcolor{paperAmber}0/40 & 1/40 \\
Qwen3-VL-32B & 40/40 & \cellcolor{paperAmber}0/40 & 0/40 \\
InternVL3.5-8B & 40/40 & \cellcolor{paperAmber}0/40 & 0/40 \\
MiniCPM-V4.5 fast & 35/40 & \cellcolor{paperAmber}0/40 & 5/40 \\
\bottomrule
\end{tabular}
\caption{Budget-paired first actions in forty held-out families per model. The highlighted branch-to-selector transition is optimal under both budgets. Arrows pair separate one- and two-observation conditions, rather than successive actions within an episode. Other pairs retain every remaining outcome. Counts are descriptive; they are not additional primary tests.}
\label{tab:open-budget-pairs}
\end{table}

%% file: open_anchor_intervals.tex
\begin{table}[!htbp]
\centering\small
\setlength{\tabcolsep}{3pt}
\begin{tabular}{@{}lrr@{}}
\toprule
Configuration & Mean loss (pp) & 95\% interval \\
\midrule
Qwen3-VL-4B & 15.62 & [14.06, 17.50] \\
Qwen3-VL-8B & 12.81 & [12.50, 13.44] \\
Qwen3-VL-32B & 12.50 & [12.50, 12.50] \\
InternVL3.5-8B & 12.50 & [12.50, 12.50] \\
MiniCPM-V4.5 fast & 13.75 & [12.81, 15.00] \\
\bottomrule
\end{tabular}
\caption{Descriptive anchor-loss intervals, resampling forty whole families with both budgets intact. Non-anchor intervals remain in Table~\ref{tab:new-open-models}. All models make zero optimal budget transitions. Under independent Bernoulli families, 0/40 has an 8.81\% two-sided 95\% Clopper--Pearson upper limit; that reference does not establish a population rate for the frozen panel. Degenerate intervals describe the finite panel.}
\label{tab:anchor-loss-intervals}
\end{table}

%% file: open_terminal_details.tex
\begin{table}[!htbp]
\centering\footnotesize
\setlength{\tabcolsep}{3pt}
\begin{tabular}{@{}lrrr@{}}
\toprule
Legal & \shortstack{Ledger $-$\\prose} & \shortstack{Ledger $-$\\active} & \shortstack{Gain\\interaction} \\
\midrule
\multicolumn{4}{@{}l}{\textbf{Qwen3-VL-4B / instruct}} \\
720/720 & \shortstack{0.42\\{}[-3.10, 3.85]} & \shortstack{5.83\\{}[1.22, 11.54]} & \shortstack{-0.50\\{}[-12.20, 12.31]} \\
\multicolumn{4}{@{}l}{\textbf{Qwen3-VL-8B / instruct}} \\
720/720 & \shortstack{0.00\\{}[-3.42, 3.10]} & \shortstack{3.75\\{}[0.21, 7.50]} & \shortstack{-3.00\\{}[-20.51, 13.49]} \\
\multicolumn{4}{@{}l}{\textbf{Qwen3-VL-32B / instruct}} \\
720/720 & \shortstack{-1.67\\{}[-4.27, 0.43]} & \shortstack{-2.08\\{}[-4.58, -0.40]} & --- \\
\multicolumn{4}{@{}l}{\textbf{InternVL3.5-8B / default}} \\
720/720 & \shortstack{-2.92\\{}[-6.84, 0.85]} & \shortstack{-3.54\\{}[-7.92, 0.85]} & --- \\
\multicolumn{4}{@{}l}{\textbf{MiniCPM-V4.5 / fast}} \\
720/720 & \shortstack{-2.92\\{}[-8.54, 2.33]} & \shortstack{-3.12\\{}[-7.92, 1.71]} & --- \\
\bottomrule
\end{tabular}
\caption{Prospective terminal confirmation under the same model configurations. Positive gain means lower answer-evidence loss with the ledger; the active-field control is separately evaluated. Gain interaction is the prose--ledger gain difference between development-diagnosis positive and negative states. All effects are percentage points with predeclared 98.75\% intervals from the four-comparison family for each model. Inference retains six states, deduplicates the shared public input and resamples 39 connected components. A dash denotes an unidentifiable contrast. $\dagger$ denotes a finite-panel degenerate interval. Diagnosis groups are locked from development separately for each model, so specificity columns need not represent identical state groups.}
\label{tab:new-open-terminal}
\end{table}

%% file: terminal_group_gains.tex
\begin{table}[!htbp]
\centering\small
\setlength{\tabcolsep}{3pt}
\begin{tabular}{@{}lrr@{}}
\toprule
Positive $n$/gain & Negative $n$/gain & Interaction \\
\midrule
\multicolumn{3}{@{}l}{\textbf{InternVL3.5-8B}} \\
0 / --- & 6 / -2.92 & --- \\
\multicolumn{3}{@{}l}{\textbf{MiniCPM-V4.5 fast}} \\
0 / --- & 6 / -2.92 & --- \\
\multicolumn{3}{@{}l}{\textbf{Qwen3-VL-32B}} \\
0 / --- & 6 / -1.67 & --- \\
\multicolumn{3}{@{}l}{\textbf{Qwen3-VL-4B}} \\
1 / 0.00 & 5 / 0.50 & -0.50 \\
\multicolumn{3}{@{}l}{\textbf{Qwen3-VL-8B}} \\
1 / -2.50 & 5 / 0.50 & -3.00 \\
\bottomrule
\end{tabular}
\caption{Development-locked group gains in declared-model expected correctness for ledger minus prose (pp). States have equal weight within each group. The difference of these gains reproduces the interaction; one diagnosis-positive state exists for Qwen-4B and Qwen-8B, while the other conditions have none. The original 39-component intervals remain in Table~\ref{tab:new-open-terminal}.}
\label{tab:terminal-group-gains}
\end{table}

%% file: core_program_references.tex
\begin{table}[!htbp]
\centering
\small
\setlength{\tabcolsep}{3pt}
\caption{Descriptive program references on the frozen acquisition panel. Loss and adaptation are in percentage points; hits are percentages. Uniform tie averaging uses every tied optimum of the stated criterion. Loss assumes exact optimal continuation after the first action; no model outputs, fitted parameters, training or compute matching are used. These references were summarized after initial model collection and do not add primary hypothesis tests.}
\label{tab:core-program-references}
\begin{tabular}{lrrrr}
\toprule
Reference & \shortstack{Non-anchor\\loss} & \shortstack{Non-anchor\\hits} & \shortstack{Anchor\\loss} & Adapt. \\
\midrule
Uniform & 3.03 & 30.00 & 17.50 & 0.00 \\
Fixed W1 & 3.03 & 30.00 & 17.50 & 0.00 \\
Menu first & 3.23 & 28.75 & 17.81 & 0.00 \\
Fixed selector & 0.60 & 50.00 & 12.50 & 0.00 \\
Fixed branch T & 2.57 & 50.00 & 12.50 & 0.00 \\
Greedy accuracy & 2.97 & 37.14 & 12.50 & -20.00 \\
Greedy entropy & 2.57 & 50.00 & 12.50 & 0.00 \\
Full horizon & 0.00 & 100.00 & 0.00 & 100.00 \\
\bottomrule
\end{tabular}
\end{table}

%% file: new_panel_methods.tex
\section{Three-Step and Natural-Form Confirmations}
\label{sec:extended-methods}
The additional protocols retain the three primary checkpoints rather than selecting a winner from core confirmation. Each planned configuration has 1,032 hierarchy requests (72 development, 960 confirmation) and 504 document requests (84 development, 420 confirmation). Confirmation requires the complete development schedule and at least 95\% strict legality. Invalid and technical outputs remain visible, and failures are not replaced or semantically repaired.

\subsection{Eight-Region Hierarchy}
The public rule has a root, two gates, four leaves and a distractor. Independent binary latent variables and declared read channels define exact finite-horizon action values. Six development families and forty confirmation families have independent role/menu assignments. Development has two fixed profiles and confirmation four; the fair profile is shared as a cross-family anchor. All first actions at budgets one, two and three are queried. At budget three, actual returned bits determine the second and third actions and final answer. The evaluator privately freezes per-region worlds and independent read-noise uniforms before requests; no unrevealed variable enters a policy prompt.

Four-profile family weighting is fixed. The root-noise profile has all eight budget-three first actions tied at .75, so its zero loss supplies no first-action discrimination. A root-first policy belongs to the optimal set in every budget-three profile. All forty families and the flat profile are retained; budget-one/two diagnostics and full trajectories constrain the interpretation. Program references replay the same latent worlds with Bayes terminal decisions. An observed-route heuristic is value-equivalent to the exact three-read reference on these profiles; realized count differences between their tie choices are not an expected-value improvement.

The two hierarchy primary endpoints are mean first-action loss at budget three and realized final correctness on all planned three-step episodes. Acquisition loss at later steps, terminal gap and budget-specific first choices are secondary diagnostics. The performance-difference identity is an expectation under the declared read model, not an equality for one realized trajectory.

\input{e4_budget_appendix}

\subsection{Profile-Specific Loss Stages and Recorded Trajectory}
\input{hierarchy_loss_intervals}
\input{hierarchy_profile_stages}
\input{recorded_trace}
\subsection{Natural-Form Inputs and Matched Readers}
We use the official FUNSD training pool for six development pages and its testing pool for thirty confirmation pages. Eligibility requires exactly one linked answer entity and a uniquely named question; semantic and transcription ambiguity are handled separately below. Stable hashes choose pages and one question per page. These choices precede experimental outputs. Each page has three input types: the native image; an unmasked overview whose longest side is resized to 512 pixels with Lanczos interpolation; and six fixed crops in three rows and two columns, numbered left to right from the top row, with ten-percent padding on each side. The candidate crop menu does not use answer boxes. Pretraining exposure to these public pages cannot be ruled out.

Each page schedules fourteen positions: native, overview and blank answers; six single-crop reader probes; a one-crop choice; two sequential choices and a dual-crop answer; and one GT-located probe. The one-crop final value is the selected cached reader response. Random and fixed crop comparators reuse those six responses, with uniform random averaged exactly, not sampled. This audit cost differs from deployment: the one-crop policy needs two calls, a cached comparator one, the two-crop policy three and native reading one. GT-located readout uses an oracle answer location but need not bound accuracy. Crop coverage is a geometric diagnostic and is not sufficient evidence of readability; the overview can itself contain the answer.

\input{document_selection_flow}
\subsection{Reference Quality and Scoring Sensitivity}
Before any document-model output, AI-assisted inspection checked all 36 pages and their original boxes; independent assistant cross-checks reviewed flagged cases. This was label-aware quality inspection rather than a human agreement study. No page, public question or candidate crop was removed or replaced. Four clear errors were corrected: neighboring-field contamination for Room and TERMS, a missing dot in CLIENT NO., and a missing word in CHAINS. Explicit literal alternatives cover low-resolution or field-boundary ambiguity in COMMENTS, STATE and a purchase-order number. Thirty-six pages (six development and thirty confirmation) have forty-one permitted references. Of 149 training and 50 test pages considered, 142 and 47 contain an eligible question; stable hashes select six and thirty. Seven and three pages respectively have no eligible question. The complete question-level exclusion audit accompanies the artifact. Literal checkbox marks remain distinct from semantic aliases, and printed ``None'' differs from JSON null.

The curated primary scorer uses NFKC, case folding and whitespace collapse, then removes only whitespace adjacent to a frozen Unicode punctuation set. This additional punctuation-space step preserves all non-whitespace characters after the base normalization. The original reference text and boxes, scored with the unchanged NFKC/casefold/whitespace exact-match rule, are applied in parallel to every saved output of every system. Coverage is the maximum union-rectangle coverage over individual accepted reference boxes. The oracle probe uses the union of those boxes; only three GT-located input images change from the archived pre-curation version, which had no model outputs. All 1,692 public policy/reader request variants are byte-identical. Formal scoring uses the recorded canonical Python 3.10.22/Unicode 13 runtime; the punctuation table is independently fixed from Unicode 15.0. The locked sensitivity changes reference text and punctuation-space handling together. A subsequent four-cell rescoring separates those factors on the same saved outputs.

\input{reference_examples}
\input{reference_flip_examples}
\subsection{Estimation and Completeness}
Each model has one additional four-endpoint family: hierarchy first loss, three-step correctness, one-crop minus uniform cached correctness, and two-crop minus native correctness. The fixed 20,000 paired bootstrap draws use seed 20261114119, resampling whole families with profiles intact or whole documents with all systems intact. We report 95\% and Bonferroni 98.75\% percentile intervals. The latter are approximate under the sampling assumptions, not finite-sample exact guarantees. Cross-model comparisons are descriptive. Non-estimable endpoints remain in the family; no failed deployment is assigned a zero confirmation effect. Original-reference sensitivity uses the complete same-output comparison, not selected favorable cases.

\subsection{Execution Coverage and Request Costs}
Tables~\ref{tab:new-panel-coverage} and~\ref{tab:new-panel-cost} report completed requests, tokens and request durations for the hierarchy and document panels.

\input{new_panel_coverage}
\input{new_panel_cost}
\subsection{Natural-Form Controls and Deployment Comparison}
Table~\ref{tab:document-controls} gives the cached-reader controls, Table~\ref{tab:deployment-comparison} the deployment-cost comparison, and Table~\ref{tab:reader-decomposition} the reader decomposition.

\input{new_document_controls}
\input{document_deployment_comparison}
\input{document_reader_decomposition}
\subsection{Scoring Contrasts and Sensitivity}
Table~\ref{tab:document-sensitivity} compares curated and original scoring, Table~\ref{tab:score-flip-counts} counts score flips on identical outputs, and Table~\ref{tab:scoring-four-cell-full} separates reference text from punctuation handling.

\input{new_document_sensitivity}
\input{score_flip_counts}
\input{scoring_factorial_appendix}
\subsection{Complete Page-Paired Outcomes}
Tables~\ref{tab:page-pairs-qwen8} and~\ref{tab:page-pairs-intern8} list every confirmation page under both references.

\input{document_page_pairs_qwen8}
\input{document_page_pairs_intern8}
\FloatBarrier

\subsection{Technical Recovery and Retained Failures}
The original InternVL single-GPU document development made 72 requests: three legal, 69 HTTP500 memory failures and 12 unreachable planned positions. Before a replacement request, an isolated source tree switched the official Qwen3 language model to its supported SDPA attention API. A small random CPU fixture checked eager/SDPA numerical agreement. The source/data analysis lock was frozen anew, with all questions, images, references, scoring, endpoints, gates and 504-request schedule unchanged. Three GPUs held the same BF16 weights without quantization. The recovery allowance was 72 additional attempts, bounding InternVL's hierarchy and both document deployments at 1,608; original failures remain separate. Recovery completed 84 legal development requests and 420 confirmation requests (418 legal, two invalid). MiniCPM document development produced 18 legal and 66 invalid outputs; no confirmation was requested. No output was semantically repaired. Attention and hardware conditions differ across deployments.

\paragraph{Data handling.}
The natural-form panel uses public FUNSD scans, which may retain names or contact details. The study collects no new human-participant data; public availability does not imply de-identification.

%% file: e4_budget_appendix.tex
\subsection{Budget-specific hierarchy diagnostics}

\label{sec:e4-budget-details}

The hierarchy contains a root, two gates, four answer leaves and one distractor. Table~\ref{tab:e4-profile-parameters} specifies the four fixed confirmation profiles. All latent variables are independent; non-root bits have prior probability .5. A read returns its true bit with the indicated correctness probability and the opposite bit otherwise. Leaf and distractor reads are noiseless. The balanced, noiseless profile is also a development anchor, while confirmation uses forty different role and menu assignments. The other development profile uses a root prior of .35, root read correctness .65 and gate read correctness .90. Profiles have equal weight.

\begin{table}[!htbp]
\centering\footnotesize
\setlength{\tabcolsep}{3pt}
\begin{tabular}{@{}lrrrrrr@{}}
\toprule
Profile & $p_r$ & $r_r$ & $r_g$ & $V_1$ & $V_2$ & $V_3$ \\
\midrule
Fair & 0.50 & 1.00 & 1.00 & 0.6250 & 0.7500 & 1.0000 \\
Flat root$^{\dagger}$ & 0.50 & 0.50 & 1.00 & 0.6250 & 0.7500 & 0.7500 \\
Root-biased & 0.20 & 0.90 & 0.95 & 0.7000 & 0.8800 & 0.9275 \\
Gate-noise & 0.70 & 0.95 & 0.70 & 0.6750 & 0.7450 & 0.8325 \\
\bottomrule
\end{tabular}
\caption{Fixed hierarchy parameters and exact initial values. $p_r$ is the root prior for bit one; $r_r$ and $r_g$ are root and gate read correctness. $V_b$ is optimal expected terminal accuracy with $b$ reads. $^{\dagger}$At three reads, all eight first actions have value .75, so this profile is a flat negative control rather than evidence of first-choice discrimination.}
\label{tab:e4-profile-parameters}
\end{table}

Table~\ref{tab:e4-budget-choices} reports each budget separately. Each cell uses all forty matched families. Hits accept every tied optimum, and loss is the conditional opportunity loss under optimal continuation. In the three non-flat profiles the root is uniquely optimal at three reads; in the negative control it is one of eight optima. Consequently, an always-root first choice is optimal in all four three-read profiles. The one- and two-read conditions and the subsequent observed histories are needed to assess budget adaptation and complete policy behavior. These are descriptive breakdowns of the existing confirmations, without additional confidence intervals or tests.

\begin{table}[!htbp]
\centering\footnotesize
\setlength{\tabcolsep}{3pt}
\begin{tabular}{@{}lrrr@{}}
\toprule
Profile & \shortstack{$b=1$\\hit / root\\loss} & \shortstack{$b=2$\\hit / root\\loss} & \shortstack{$b=3$\\hit / root\\loss} \\
\midrule
\multicolumn{4}{@{}l}{\textbf{Qwen3-VL-8B}} \\
Fair & \shortstack{2.5 / 92.5\\12.19} & \shortstack{97.5 / 92.5\\0.31} & \shortstack{92.5 / 92.5\\1.25} \\
Flat root$^{\dagger}$ & \shortstack{2.5 / 92.5\\12.19} & \shortstack{5.0 / 95.0\\11.88} & \shortstack{100.0 / 95.0\\0.00} \\
Biased & \shortstack{7.5 / 82.5\\18.25} & \shortstack{2.5 / 87.5\\15.36} & \shortstack{90.0 / 90.0\\0.46} \\
Gate-noise & \shortstack{5.0 / 85.0\\16.25} & \shortstack{2.5 / 87.5\\1.36} & \shortstack{90.0 / 90.0\\0.73} \\
\multicolumn{4}{@{}l}{\textbf{InternVL3.5-8B}} \\
Fair & \shortstack{2.5 / 82.5\\12.19} & \shortstack{95.0 / 80.0\\0.62} & \shortstack{80.0 / 80.0\\3.12} \\
Flat root$^{\dagger}$ & \shortstack{0.0 / 92.5\\12.50} & \shortstack{10.0 / 90.0\\11.25} & \shortstack{100.0 / 90.0\\0.00} \\
Biased & \shortstack{2.5 / 90.0\\19.50} & \shortstack{5.0 / 90.0\\14.85} & \shortstack{90.0 / 90.0\\0.44} \\
Gate-noise & \shortstack{0.0 / 95.0\\17.31} & \shortstack{2.5 / 95.0\\0.89} & \shortstack{95.0 / 95.0\\0.29} \\
\multicolumn{4}{@{}l}{\textbf{MiniCPM-V4.5 (fast)}} \\
Fair & \shortstack{0.0 / 100.0\\12.50} & \shortstack{100.0 / 100.0\\0.00} & \shortstack{100.0 / 100.0\\0.00} \\
Flat root$^{\dagger}$ & \shortstack{0.0 / 95.0\\12.50} & \shortstack{7.5 / 92.5\\11.56} & \shortstack{100.0 / 92.5\\0.00} \\
Biased & \shortstack{0.0 / 95.0\\20.00} & \shortstack{0.0 / 92.5\\15.69} & \shortstack{92.5 / 92.5\\0.36} \\
Gate-noise & \shortstack{0.0 / 95.0\\17.31} & \shortstack{2.5 / 95.0\\0.89} & \shortstack{92.5 / 92.5\\0.36} \\
\bottomrule
\end{tabular}
\caption{Budget-specific first choices on the hierarchy. Cells show optimal-action hits in percent / root-first choices in percent / mean expected-accuracy loss in percentage points. Every model--profile--budget cell contains forty families. $^{\dagger}$The three-read hit rate is necessarily 100\% for any legal action in the flat negative control; its zero loss does not demonstrate planning.}
\label{tab:e4-budget-choices}
\end{table}

Table~\ref{tab:e4-program-references} replays eight program policies on the same 160 frozen worlds (forty families and four profiles). All use the known observation model and a Bayes terminal decoder, choosing zero at a posterior tie. Fixed menu order reads the first remaining menu item; fixed role order reads the root followed by the two gates. The observed-route rule reads the root, the gate indicated by its return and the leaf indicated by that gate. Myopic accuracy and entropy score one observation; depth-two lookahead caps the search at two remaining reads; the remaining-horizon reference uses the full budget. Criterion ties follow menu order.

\begin{table}[!htbp]
\centering\footnotesize
\setlength{\tabcolsep}{3pt}
\begin{tabular}{@{}lrrr@{}}
\toprule
Program policy & Correct & \shortstack{First\\loss} & \shortstack{Total\\loss} \\
\midrule
Fixed menu order & 100/160 & 7.43 & 23.47 \\
Fixed role order & 69/160 & 0.00 & 37.42 \\
Observed-route rule & 138/160 & 0.00 & 0.00 \\
Seeded uniform choice & 98/160 & 7.32 & 23.03 \\
Myopic accuracy & 117/160 & 9.63 & 16.05 \\
Myopic entropy & 123/160 & 9.63 & 10.33 \\
Depth-two lookahead & 130/160 & 3.76 & 3.76 \\
Full horizon & 140/160 & 0.00 & 0.00 \\
\bottomrule
\end{tabular}
\caption{Descriptive program-policy replay at three reads. Correctness is realized on the fixed sampled worlds; the loss columns average conditional opportunity losses, with cumulative loss summing all three acquisitions. The Bayes decoder contributes zero terminal answer-evidence gap. These known-model references are neither learned policies nor compute-matched VLM systems.}
\label{tab:e4-program-references}
\end{table}

An existing exhaustive return-path calculation establishes that the observed-route rule and the remaining-horizon reference have the same expected value on these four three-read profiles. Their realized totals, 138/160 and 140/160, therefore do not establish an expected-value advantage; different optimal tie choices can yield different outcomes on the sampled worlds. The seeded uniform policy uses one fixed-seed trajectory per world. Its 98/160 correct answers are a realization of that replay, not the exact expectation of uniform random choice. The performance-difference accounting holds under the declared generative expectation, rather than equating a conditional-loss sum with each world's realized correctness difference.

%% file: hierarchy_loss_intervals.tex
\begin{table}[!htbp]
\centering\footnotesize
\setlength{\tabcolsep}{2pt}
\begin{tabular}{@{}lrrr@{}}
\toprule
Stage & Qwen-8B & InternVL & MiniCPM \\
\midrule
First (pp) & \shortstack{0.61\\{}[0.06, 1.39]} & \shortstack{0.96\\{}[0.26, 1.85]} & \shortstack{0.18\\{}[0.00, 0.47]} \\
Later (pp) & \shortstack{24.52\\{}[22.26, 26.70]} & \shortstack{22.20\\{}[20.33, 24.12]} & \shortstack{26.76\\{}[24.02, 29.46]} \\
Final (pp) & \shortstack{5.73\\{}[3.61, 7.96]} & \shortstack{14.30\\{}[10.23, 18.92]} & \shortstack{8.79\\{}[5.55, 12.22]} \\
Best (\%) & 62.43 & 63.98 & 60.71 \\
Actual (\%) & \shortstack{53.75\\{}[43.75, 63.13]} & \shortstack{57.50\\{}[46.87, 68.12]} & \shortstack{57.50\\{}[46.25, 68.12]} \\
\bottomrule
\end{tabular}
\caption{Complete three-read loss and accuracy intervals for Qwen3-VL-8B, InternVL3.5-8B and MiniCPM-V4.5 fast. Each has 160 histories in forty families. First loss and actual accuracy use 98.75\% family intervals; later and final loss use descriptive 95\% intervals (20,000 draws). Best is mean $V_0(H_B)$; final is model-relative terminal loss.}
\label{tab:hierarchy-loss-intervals}
\end{table}

%% file: hierarchy_profile_stages.tex
\begin{table}[!htbp]
\centering\footnotesize
\setlength{\tabcolsep}{3pt}
\begin{tabular}{@{}lrrrr@{}}
\toprule
Profile (losses in pp) & First & Later & Final & $V_0$ (\%) \\
\midrule
\multicolumn{5}{@{}l}{\textbf{Qwen3-VL-8B}} \\
biased & 0.46 & 31.26 & 2.21 & 61.39 \\
fair & 1.25 & 35.62 & 5.62 & 61.56 \\
gate noise & 0.73 & 16.82 & 11.96 & 65.51 \\
root noise & 0.00 & 14.37 & 3.12 & 61.25 \\
\multicolumn{5}{@{}l}{\textbf{InternVL3.5-8B}} \\
biased & 0.44 & 28.37 & 20.11 & 64.14 \\
fair & 3.12 & 29.38 & 11.88 & 65.62 \\
gate noise & 0.29 & 17.62 & 15.85 & 65.51 \\
root noise & 0.00 & 13.44 & 9.38 & 60.62 \\
\multicolumn{5}{@{}l}{\textbf{MiniCPM-V4.5 fast}} \\
biased & 0.36 & 35.04 & 7.10 & 56.33 \\
fair & 0.00 & 36.88 & 8.75 & 62.50 \\
gate noise & 0.36 & 20.75 & 10.56 & 63.39 \\
root noise & 0.00 & 14.37 & 8.75 & 60.62 \\
\bottomrule
\end{tabular}
\caption{Loss stages before averaging the four profiles. Each row contains forty actual three-read histories.}
\label{tab:hierarchy-profile-stages}
\end{table}

%% file: recorded_trace.tex
\begin{table}[!htbp]
\centering\small
\setlength{\tabcolsep}{3pt}
\begin{tabular}{@{}lrrr@{}}
\toprule
Step & Choice & Return & Loss (pp) \\
\midrule
Read 1 & Root W6 & 0 & 0.0 \\
Read 2 & Leaf W7 & 1 & \cellcolor{paperAmber}25.0 \\
Read 3 & Leaf W3 & 0 & 0.0 \\
\midrule
Answer & \shortstack{0\\truth: 1} & \shortstack{$V_0=0.5$\\$\eta=0$} & --- \\
\bottomrule
\end{tabular}
\caption{A recorded Qwen3-VL-8B fair-profile bit-return rollout. After W6 returns zero, W4 is the unresolved routing gate. The second choice opens W7 instead. The final posterior is tied despite an incorrect realized answer. Selection uses the first qualifying trace in fixed order, rather than the largest error.}
\label{tab:recorded-trace}
\end{table}

%% file: document_selection_flow.tex
\begin{table}[!htbp]
\centering\small
\setlength{\tabcolsep}{3pt}
\begin{tabular}{@{}lrr@{}}
\toprule
Source / study split & \shortstack{All /\\eligible} & \shortstack{Excluded /\\selected} \\
\midrule
Training / dev. & 149 / 142 & 7 / 6 \\
Test / conf. & 50 / 47 & 3 / 30 \\
\bottomrule
\end{tabular}
\caption{Frozen page selection before inference. Eligible questions have one linked answer, unique field text and satisfy the recorded text and geometry checks. Stable hashes select one eligible question per page and the study subset; no model output filters pages. The artifact retains question-level exclusion counts and the source-to-study mapping.}
\label{tab:document-selection-flow}
\end{table}

%% file: reference_examples.tex
\begin{table}[!htbp]
\centering\small
\setlength{\tabcolsep}{3pt}
\begin{tabular}{@{}p{.21\linewidth}p{.72\linewidth}@{}}
\toprule
Field / item & Literal value or interpretation \\
\midrule
\multicolumn{2}{@{}l}{\textbf{Room}} \\
Original & 803E // Pages (including cover) \\
Revised & 803E \\
Reason & Adjacent field included. \\
\multicolumn{2}{@{}l}{\textbf{TERMS}} \\
Original & 27407 NET 15 F. O. B. N/A VIA N/A \\
Revised & NET 15 \\
Reason & Neighboring entries included. \\
\multicolumn{2}{@{}l}{\textbf{CLIENT NO.}} \\
Original & L8557 002 \\
Revised & L8557.002 \\
Reason & Visible separator dot omitted. \\
\bottomrule
\end{tabular}
\caption{Examples of reference corrections fixed before document-model outputs. These literal revisions are distinct from punctuation-adjacent whitespace normalization. Ambiguous fields retain multiple predeclared literal references; no semantic aliases are introduced.}
\label{tab:reference-examples}
\end{table}

%% file: reference_flip_examples.tex
\begin{table}[!htbp]
\centering\small
\setlength{\tabcolsep}{3pt}
\begin{tabular}{@{}p{.21\linewidth}p{.72\linewidth}@{}}
\toprule
Field / item & Literal value or interpretation \\
\midrule
\multicolumn{2}{@{}l}{\textbf{Room}} \\
Original & 803E // Pages (including cover) \\
Revised & 803E \\
Saved output & 803E \\
Score O$\to$C & 0$\to$1 \\
\multicolumn{2}{@{}l}{\textbf{TERMS}} \\
Original & 27407 NET 15 F. O. B. N/A VIA N/A \\
Revised & NET 15 \\
Saved output & NET 15 \\
Score O$\to$C & 0$\to$1 \\
\multicolumn{2}{@{}l}{\textbf{CLIENT NO.}} \\
Original & L8557 002 \\
Revised & L8557.002 \\
Saved output & L8557.002 \\
Score O$\to$C & 0$\to$1 \\
\bottomrule
\end{tabular}
\caption{Illustrative, non-adjudicated Qwen3-VL-8B native outputs under the two frozen scoring definitions. Reference revisions precede model outputs. Room and TERMS remove neighboring fields; CLIENT NO. restores a visible dot. Full page-paired flips are in the appendix.}
\label{tab:reference-flip-examples}
\end{table}

%% file: new_panel_coverage.tex
\begin{table}[!htbp]
\centering\footnotesize
\setlength{\tabcolsep}{3pt}
\begin{tabular}{@{}lrrr@{}}
\toprule
Task / split & Planned & Calls & Legal \\
\midrule
\multicolumn{4}{@{}l}{\textbf{Qwen3-VL-8B}} \\
Hierarchy / dev. & 72 & 72 & 72 \\
Hierarchy / conf. & 960 & 960 & 960 \\
Forms / dev. & 84 & 84 & 84 \\
Forms / conf. & 420 & 420 & 420 \\
\multicolumn{4}{@{}l}{\textbf{InternVL3.5-8B}} \\
Hierarchy / dev. & 72 & 72 & 72 \\
Hierarchy / conf. & 960 & 960 & 960 \\
Forms / dev. & 84 & 72 & 3 \\
\multicolumn{4}{@{}l}{\textbf{InternVL3.5-8B (SDPA)}} \\
Forms / dev. & 84 & 84 & 84 \\
Forms / conf. & 420 & 420 & 418 \\
\multicolumn{4}{@{}l}{\textbf{MiniCPM-V4.5 fast}} \\
Hierarchy / dev. & 72 & 72 & 72 \\
Hierarchy / conf. & 960 & 960 & 960 \\
Forms / dev. & 84 & 84 & 18 \\
\bottomrule
\end{tabular}
\caption{New-panel execution coverage. Scheduled steps count planned positions; model requests count actual API attempts. InternVL's original single-GPU development has 12 unreachable positions after failed actions; these consume no request. All failures and any separately labeled recovery remain visible. Confirmation requires the complete development schedule with at least 95\% strict legality.}
\label{tab:new-panel-coverage}
\end{table}

%% file: new_panel_cost.tex
\begin{table}[!htbp]
\centering\footnotesize
\setlength{\tabcolsep}{3pt}
\begin{tabular}{@{}lrrrr@{}}
\toprule
Task / split & Usage / calls & In tokens & Out tokens & Seconds \\
\midrule
\multicolumn{5}{@{}l}{\textbf{Qwen3-VL-8B}} \\
Hierarchy / dev. & 72/72 & 27,264 & 492 & 29.8 \\
Hierarchy / conf. & 960/960 & 363,520 & 6,560 & 366.8 \\
Forms / dev. & 84/84 & 38,781 & 1,038 & 61.0 \\
Forms / conf. & 420/420 & 192,057 & 4,594 & 264.4 \\
\multicolumn{5}{@{}l}{\textbf{InternVL3.5-8B}} \\
Hierarchy / dev. & 72/72 & 29,928 & 492 & 32.6 \\
Hierarchy / conf. & 960/960 & 399,040 & 6,560 & 407.7 \\
Forms / dev. & 3/72 & 10,321 & 24 & 32.6 \\
\multicolumn{5}{@{}l}{\textbf{InternVL3.5-8B (SDPA)}} \\
Forms / dev. & 84/84 & 344,640 & 892 & 184.7 \\
Forms / conf. & 420/420 & 1,730,372 & 4,784 & 922.7 \\
\multicolumn{5}{@{}l}{\textbf{MiniCPM-V4.5 fast}} \\
Hierarchy / dev. & 72/72 & 27,552 & 492 & 28.0 \\
Hierarchy / conf. & 960/960 & 367,360 & 6,560 & 357.3 \\
Forms / dev. & 84/84 & 25,689 & 3,734 & 177.2 \\
\bottomrule
\end{tabular}
\caption{Actual new-panel request costs, including failed requests where measured. Usage / requests gives the number of requests with service-reported token usage over all attempts. The original InternVL forms condition has token measurements for only 3/72 attempts; its token total is incomplete. The SDPA forms condition uses three GPUs and the original eager condition one; other deployment details are in the artifact guide. Usage is service-reported; request seconds are summed client durations, not concurrent wall-clock latency. Original deployment failures and recovery are separate conditions.}
\label{tab:new-panel-cost}
\end{table}

%% file: new_document_controls.tex
\begin{table}[!htbp]
\centering\small
\setlength{\tabcolsep}{3pt}
\begin{tabular}{@{}lrr@{}}
\toprule
System / diagnostic & \shortstack{Qwen\\C / O} & \shortstack{InternVL\\C / O} \\
\midrule
Whole native & 83.33 / 50.00 & 83.33 / 53.33 \\
Overview & 46.67 / 33.33 & 70.00 / 46.67 \\
Blank & 0.00 / 0.00 & 0.00 / 0.00 \\
Uniform cached & 55.56 / 39.44 & 68.33 / 44.44 \\
Policy b1 & 80.00 / 50.00 & 73.33 / 46.67 \\
Policy b2 & 70.00 / 50.00 & 73.33 / 46.67 \\
GT-located reader & 86.67 / 53.33 & 80.00 / 50.00 \\
Fixed R1 & 60.00 / 40.00 & 70.00 / 43.33 \\
Fixed R2 & 63.33 / 40.00 & 70.00 / 46.67 \\
Fixed R3 & 63.33 / 46.67 & 66.67 / 43.33 \\
Fixed R4 & 43.33 / 33.33 & 66.67 / 43.33 \\
Fixed R5 & 56.67 / 40.00 & 70.00 / 46.67 \\
Fixed R6 & 46.67 / 36.67 & 66.67 / 43.33 \\
b1 coverage & 63.33 / 56.67 & 50.00 / 46.67 \\
b2 coverage & 63.33 / 60.00 & 66.67 / 66.67 \\
\bottomrule
\end{tabular}
\caption{All natural-form controls and reader probes on the same 30 confirmation documents, in percent. Uniform cached is the exact average over six reader probes. Policy b1 uses two deployment calls; b2 uses three, whole native one. GT-located is an oracle-location readout reference, not an accuracy upper bound. Coverage uses the selected rectangles and each accepted reference box (maximum coverage across variants); it is not readability.}
\label{tab:document-controls}
\end{table}

%% file: document_deployment_comparison.tex
\begin{table}[!htbp]
\centering\small
\setlength{\tabcolsep}{3pt}
\begin{tabular}{@{}lrrr@{}}
\toprule
System & Calls & Input tokens & Mpixel \\
\midrule
\multicolumn{4}{@{}l}{\textbf{Qwen3-VL-8B}} \\
Whole native & 1.00 & 824.8 & 0.766 \\
Overview & 1.00 & 269.3 & 0.201 \\
One crop & 2.00 & 901.2 & 0.561 \\
Two crops & 3.00 & 1662.8 & 1.082 \\
\multicolumn{4}{@{}l}{\textbf{InternVL3.5-8B}} \\
Whole native & 1.00 & 3441.6 & 0.766 \\
Overview & 1.00 & 3441.6 & 0.201 \\
One crop & 2.00 & 8070.0 & 0.557 \\
Two crops & 3.00 & 13242.8 & 1.079 \\
\bottomrule
\end{tabular}
\caption{Observed per-page deployment costs reconstructed from saved requests; the one-crop reader is reused from its frozen probe. Decoded pixel area sums all images across calls, counting repeated overview inputs. It is not vision-token count or an equal-pixel control. InternVL uses its recovered SDPA deployment; architecture, inputs and compute are not causally isolated.}
\label{tab:deployment-comparison}
\end{table}

%% file: document_reader_decomposition.tex
\begin{table}[!htbp]
\centering\footnotesize
\setlength{\tabcolsep}{3pt}
\begin{tabular}{@{}lrr@{}}
\toprule
Model / reference & Covered & Correct: in / out \\
\midrule
Qwen3-VL-8B / Curated & 19/30 & 17/19 / 7/11 \\
Qwen3-VL-8B / Original & 17/30 & 11/17 / 4/13 \\
InternVL3.5-8B / Curated & 15/30 & 13/15 / 9/15 \\
InternVL3.5-8B / Original & 14/30 & 8/14 / 6/16 \\
\bottomrule
\end{tabular}
\caption{One-crop outcomes conditional on geometric coverage. Covered means at least the frozen 95\% reference-box coverage threshold. This is a location diagnostic, not optimal crop selection or readability. The final one-crop answer is the selected cached reader output, so reader and final correctness are identical by design; they cannot be independently attributed here.}
\label{tab:reader-decomposition}
\end{table}

%% file: new_document_sensitivity.tex
\begin{table}[!htbp]
\centering\footnotesize
\setlength{\tabcolsep}{3pt}
\begin{tabular}{@{}lrr@{}}
\toprule
Scoring & Contrast & Estimate [98.75\% CI] \\
\midrule
\multicolumn{3}{@{}l}{\textbf{Qwen3-VL-8B}} \\
Curated & 1 crop $-$ uniform & \shortstack{24.44\\{}[8.89, 41.11]} \\
Curated & 2 crops $-$ native & \shortstack{-13.33\\{}[-33.33, 6.67]} \\
Original & 1 crop $-$ uniform & \shortstack{10.56\\{}[-0.56, 24.44]} \\
Original & 2 crops $-$ native & \shortstack{0.00\\{}[-13.33, 13.33]} \\
\multicolumn{3}{@{}l}{\textbf{InternVL3.5-8B}} \\
Curated & 1 crop $-$ uniform & \shortstack{5.00\\{}[-1.67, 16.67]} \\
Curated & 2 crops $-$ native & \shortstack{-10.00\\{}[-26.67, 0.00]} \\
Original & 1 crop $-$ uniform & \shortstack{2.22\\{}[-2.22, 11.11]} \\
Original & 2 crops $-$ native & \shortstack{-6.67\\{}[-20.00, 0.00]} \\
\bottomrule
\end{tabular}
\caption{Complete original-reference scoring sensitivity on the same saved model outputs. The primary curated references were frozen before experimental outputs. The original-reference exact score retains the original NFKC/casefold/whitespace rule and original answer text; this is a measurement sensitivity analysis, not an annotation intervention experiment.}
\label{tab:document-sensitivity}
\end{table}

%% file: score_flip_counts.tex
\begin{table}[!htbp]
\centering\small
\setlength{\tabcolsep}{3pt}
\begin{tabular}{@{}lrrr@{}}
\toprule
System & 0 to 1 & 1 to 0 & Same \\
\midrule
\multicolumn{4}{@{}l}{\textbf{Qwen3-VL-8B}} \\
Native & 10 & 0 & 20 \\
One crop & 9 & 0 & 21 \\
Two crops & 6 & 0 & 24 \\
\multicolumn{4}{@{}l}{\textbf{InternVL3.5-8B}} \\
Native & 9 & 0 & 21 \\
One crop & 8 & 0 & 22 \\
Two crops & 8 & 0 & 22 \\
\bottomrule
\end{tabular}
\caption{Paired scoring flips from original to curated definitions on all thirty documents. Changes combine reference text and punctuation-space normalization.}
\label{tab:score-flip-counts}
\end{table}

%% file: scoring_factorial_appendix.tex
\begin{table}[!htbp]
\centering\footnotesize
\setlength{\tabcolsep}{3pt}
\begin{tabular}{@{}lrr@{}}
\toprule
Score & Contrast / accuracy & Estimate [98.75\% CI] \\
\midrule
\multicolumn{3}{@{}l}{\textbf{Qwen3-VL-8B}} \\
OO & Selection gain & \shortstack{10.56\\{}[-0.56, 24.44]} \\
OO & System contrast & \shortstack{0.00\\{}[-13.33, 13.33]} \\
OO & Six-crop oracle & \shortstack{56.67\\{}[33.33, 80.00]} \\
CO & Selection gain & \shortstack{22.78\\{}[7.78, 39.44]} \\
CO & System contrast & \shortstack{-13.33\\{}[-33.33, 6.67]} \\
CO & Six-crop oracle & \shortstack{83.33\\{}[63.33, 96.67]} \\
OC & Selection gain & \shortstack{20.56\\{}[6.11, 36.67]} \\
OC & System contrast & \shortstack{-10.00\\{}[-30.00, 6.67]} \\
OC & Six-crop oracle & \shortstack{70.00\\{}[49.98, 90.00]} \\
CC & Selection gain & \shortstack{24.44\\{}[8.89, 41.11]} \\
CC & System contrast & \shortstack{-13.33\\{}[-33.33, 6.67]} \\
CC & Six-crop oracle & \shortstack{86.67\\{}[70.00, 100.00]} \\
\multicolumn{3}{@{}l}{\textbf{InternVL3.5-8B}} \\
OO & Selection gain & \shortstack{2.22\\{}[-2.22, 11.11]} \\
OO & System contrast & \shortstack{-6.67\\{}[-20.00, 0.00]} \\
OO & Six-crop oracle & \shortstack{50.00\\{}[26.67, 73.33]} \\
CO & Selection gain & \shortstack{5.00\\{}[-1.67, 16.67]} \\
CO & System contrast & \shortstack{-10.00\\{}[-26.67, 0.00]} \\
CO & Six-crop oracle & \shortstack{76.67\\{}[56.67, 93.33]} \\
OC & Selection gain & \shortstack{2.22\\{}[-2.22, 11.11]} \\
OC & System contrast & \shortstack{-6.67\\{}[-20.00, 0.00]} \\
OC & Six-crop oracle & \shortstack{63.33\\{}[40.00, 83.33]} \\
CC & Selection gain & \shortstack{5.00\\{}[-1.67, 16.67]} \\
CC & System contrast & \shortstack{-10.00\\{}[-26.67, 0.00]} \\
CC & Six-crop oracle & \shortstack{76.67\\{}[56.67, 93.33]} \\
\bottomrule
\end{tabular}
\caption{Post-hoc four-cell scoring of the same saved outputs on thirty pages. The first letter gives original (O) or revised (C) reference text; the second gives base (O) or punctuation-space (C) normalization. Intervals use the locked document-pair resampling convention (98.75\%); the original and revised diagonals reproduce existing results. New off-diagonal cells and the six-crop oracle are descriptive, without new confirmatory tests. The oracle answers correctly whenever any of the six cached readers is correct and is not an implementable selection policy. All recorded scoring booleans are reproduced before generating the off-diagonal cells.}
\label{tab:scoring-four-cell-full}
\end{table}

%% file: document_page_pairs_qwen8.tex
\begin{table}[!htbp]
\centering\footnotesize
\setlength{\tabcolsep}{3pt}
\begin{tabular}{@{}lrrrr@{}}
\toprule
Page & \shortstack{Native\\C / O} & \shortstack{1 crop\\C / O} & \shortstack{2 crops\\C / O} & \shortstack{Uniform\\C / O} \\
\midrule
1 & 0 / 0 & 0 / 0 & 0 / 0 & 0/6 / 0/6 \\
2 & 1 / 1 & 1 / 1 & 1 / 1 & 6/6 / 6/6 \\
3 & 1 / 0 & 1 / 0 & 1 / 0 & 4/6 / 0/6 \\
4 & 1 / 1 & 1 / 1 & 1 / 1 & 2/6 / 2/6 \\
5 & 1 / 1 & 1 / 1 & 1 / 1 & 2/6 / 2/6 \\
6 & 1 / 1 & 1 / 1 & 1 / 1 & 6/6 / 6/6 \\
7 & 1 / 1 & 1 / 1 & 1 / 1 & 6/6 / 6/6 \\
8 & 1 / 1 & 1 / 1 & 1 / 1 & 1/6 / 1/6 \\
9 & 1 / 1 & 1 / 1 & 1 / 1 & 6/6 / 6/6 \\
10 & 1 / 1 & 1 / 1 & 1 / 1 & 5/6 / 5/6 \\
11 & 1 / 1 & 0 / 0 & 0 / 0 & 1/6 / 1/6 \\
12 & 1 / 0 & 1 / 0 & 1 / 0 & 1/6 / 0/6 \\
13 & 1 / 0 & 1 / 0 & 1 / 0 & 3/6 / 2/6 \\
14 & 1 / 1 & 1 / 1 & 1 / 1 & 6/6 / 6/6 \\
15 & 1 / 0 & 1 / 0 & 0 / 0 & 2/6 / 0/6 \\
16 & 1 / 0 & 1 / 0 & 1 / 0 & 1/6 / 0/6 \\
17 & 1 / 1 & 1 / 1 & 1 / 1 & 2/6 / 2/6 \\
18 & 1 / 1 & 1 / 1 & 1 / 1 & 6/6 / 6/6 \\
19 & 1 / 1 & 1 / 1 & 1 / 1 & 6/6 / 6/6 \\
20 & 1 / 0 & 1 / 0 & 0 / 0 & 6/6 / 0/6 \\
21 & 1 / 0 & 1 / 0 & 1 / 0 & 6/6 / 0/6 \\
22 & 0 / 0 & 0 / 0 & 0 / 0 & 0/6 / 0/6 \\
23 & 1 / 0 & 1 / 0 & 1 / 0 & 6/6 / 0/6 \\
24 & 0 / 0 & 0 / 0 & 0 / 0 & 0/6 / 0/6 \\
25 & 1 / 1 & 1 / 1 & 1 / 1 & 6/6 / 6/6 \\
26 & 0 / 0 & 1 / 1 & 1 / 1 & 2/6 / 2/6 \\
27 & 1 / 0 & 0 / 0 & 0 / 0 & 1/6 / 0/6 \\
28 & 1 / 1 & 1 / 1 & 1 / 1 & 6/6 / 6/6 \\
29 & 1 / 0 & 1 / 0 & 0 / 0 & 1/6 / 0/6 \\
30 & 0 / 0 & 0 / 0 & 0 / 0 & 0/6 / 0/6 \\
\bottomrule
\end{tabular}
\caption{Complete paired confirmation outcomes for Qwen3-VL-8B. C and O use curated and original-reference exact scoring on the same saved outputs. Native and policy entries are binary correctness; uniform entries count correct cached readers out of six. Page indices follow the frozen evaluation order; their source mapping accompanies the artifact.}
\label{tab:page-pairs-qwen8}
\end{table}

%% file: document_page_pairs_intern8.tex
\begin{table}[!htbp]
\centering\footnotesize
\setlength{\tabcolsep}{3pt}
\begin{tabular}{@{}lrrrr@{}}
\toprule
Page & \shortstack{Native\\C / O} & \shortstack{1 crop\\C / O} & \shortstack{2 crops\\C / O} & \shortstack{Uniform\\C / O} \\
\midrule
1 & 0 / 0 & 0 / 0 & 0 / 0 & 0/6 / 0/6 \\
2 & 1 / 1 & 1 / 1 & 1 / 1 & 6/6 / 6/6 \\
3 & 1 / 0 & 1 / 0 & 1 / 0 & 6/6 / 0/6 \\
4 & 1 / 1 & 1 / 1 & 1 / 1 & 6/6 / 6/6 \\
5 & 1 / 1 & 1 / 1 & 1 / 1 & 1/6 / 1/6 \\
6 & 1 / 1 & 1 / 1 & 1 / 1 & 6/6 / 6/6 \\
7 & 1 / 1 & 1 / 1 & 1 / 1 & 6/6 / 6/6 \\
8 & 1 / 1 & 0 / 0 & 0 / 0 & 1/6 / 1/6 \\
9 & 1 / 1 & 1 / 1 & 1 / 1 & 6/6 / 6/6 \\
10 & 1 / 1 & 1 / 1 & 1 / 1 & 6/6 / 6/6 \\
11 & 1 / 1 & 1 / 1 & 1 / 1 & 6/6 / 6/6 \\
12 & 1 / 0 & 1 / 0 & 1 / 0 & 1/6 / 0/6 \\
13 & 1 / 0 & 1 / 0 & 1 / 0 & 6/6 / 0/6 \\
14 & 1 / 1 & 1 / 1 & 1 / 1 & 6/6 / 6/6 \\
15 & 1 / 0 & 1 / 0 & 1 / 0 & 6/6 / 0/6 \\
16 & 1 / 0 & 1 / 0 & 1 / 0 & 6/6 / 0/6 \\
17 & 1 / 1 & 1 / 1 & 1 / 1 & 6/6 / 6/6 \\
18 & 1 / 1 & 1 / 1 & 1 / 1 & 6/6 / 6/6 \\
19 & 1 / 1 & 1 / 1 & 1 / 1 & 6/6 / 6/6 \\
20 & 1 / 0 & 1 / 0 & 1 / 0 & 6/6 / 0/6 \\
21 & 1 / 0 & 1 / 0 & 1 / 0 & 6/6 / 0/6 \\
22 & 0 / 0 & 0 / 0 & 0 / 0 & 0/6 / 0/6 \\
23 & 1 / 0 & 0 / 0 & 0 / 0 & 0/6 / 0/6 \\
24 & 0 / 0 & 0 / 0 & 0 / 0 & 0/6 / 0/6 \\
25 & 1 / 1 & 1 / 1 & 1 / 1 & 6/6 / 6/6 \\
26 & 0 / 0 & 0 / 0 & 0 / 0 & 0/6 / 0/6 \\
27 & 0 / 0 & 0 / 0 & 0 / 0 & 0/6 / 0/6 \\
28 & 1 / 1 & 1 / 1 & 1 / 1 & 6/6 / 6/6 \\
29 & 1 / 0 & 1 / 0 & 1 / 0 & 6/6 / 0/6 \\
30 & 1 / 1 & 0 / 0 & 0 / 0 & 0/6 / 0/6 \\
\bottomrule
\end{tabular}
\caption{Complete paired confirmation outcomes for InternVL3.5-8B. C and O use curated and original-reference exact scoring on the same saved outputs. Native and policy entries are binary correctness; uniform entries count correct cached readers out of six. Page indices follow the frozen evaluation order; their source mapping accompanies the artifact.}
\label{tab:page-pairs-intern8}
\end{table}

%% file: diagnostic_accounting.tex
\section{Additional Task-Model Accounting}
\label{sec:appendix-accounting}

\paragraph{Proof of policy return accounting.}
Under the information and channel assumptions in Sec.~\ref{sec:formal-model}, define $\delta_t=\delta_{B-t}(H_t,A_t)$ and $\eta_B=\eta(H_B,\hat Y)$. Every expectation in this paragraph conditions on $(C,H_0)$ and integrates over the prior, read noise and both policies. By the action-value recursion and the tower property,
\begin{equation}
\begin{aligned}
 \mathbb E[\delta_t]
 &=\mathbb E[V_{B-t}(H_t)-Q_{B-t}(H_t,A_t)]\\
 &=\mathbb E[V_{B-t}(H_t)-V_{B-t-1}(H_{t+1})].
\end{aligned}
\label{eq:one-step-telescope}
\end{equation}
Summing over $t=0,\ldots,B-1$ cancels the intermediate values. The terminal information restriction gives $\mathbb E[\eta_B]=\mathbb E[V_0(H_B)]-J_\mu$. Combining these equalities yields
\begin{equation}
\begin{aligned}
 V_B(H_0)-J_\mu
 &=\mathbb E\!\left[\sum_{t=0}^{B-1}\delta_t+\eta_B\right].
\end{aligned}
\label{eq:trajectory-loss}
\end{equation}
Each loss is nonnegative by maximization. The identity concerns expectation under the declared generative model; it does not equate a conditional-value sum with realized correctness on each world. Resampling observed families estimates sampling uncertainty on the fixed panel, not correctness of the model assumptions. Restricting to completed legal traces can select a different history distribution, so their mean losses alone do not establish this identity for a policy with invalid outputs.

This accounting suggests different repairs. On fixed acquired histories, choosing the posterior mode removes $\eta_B$ in the task model. The expected sum of $\delta_t$ measures room for better acquisition against that terminal reference. Replacing the acquisition policy can change the histories and thus change the errors of the original answer model. Acquisition loss alone does not predict that end-to-end gain. The archived inactive-branch reads and the wrong answer after decisive evidence fall on different sides of Eq.~\ref{eq:trajectory-loss}. The ledger's three corrected answers on identical crop histories provide a direct example of a terminal intervention. When the loss occurs before the evidence arrives, as in the moving-boundary decisions, a planner that scores candidate observations targets a different step.

\paragraph{Value a routing read by what it enables.}
Suppose the first read returns $z$, and $g_a(z)$ is the expected terminal accuracy after a follow-up $a$. For the same set of available follow-ups, the value of selecting one after seeing $z$, relative to committing before that read, is
\begin{equation}
  \mathbb{E}_z[\max_a g_a(z)]
     -\max_a\mathbb{E}_z[g_a(z)]\ \geq\ 0.
  \label{eq:routing-option}
\end{equation}
In the fair switch task the selector leaves the answer posterior tied. Reading it and then a fixed branch gives .75 expected accuracy, whereas routing the second read to the active branch gives 1.00. The .25 difference comes from conditioning the next action on the first return. A score based on the first read's immediate answer gain cannot assign it that value. With fair bits and symmetric read errors, selector reliability $\rho\geq .5$ gives the indicated branch posterior routing weight $\rho$, versus .5 for a fixed branch on average. If either branch has read reliability $r\geq .5$, its read improves accuracy by $(r-.5)$ times that weight. The adaptive second read has expected accuracy $.5+\rho(r-.5)$, while committing to either fixed branch has accuracy $.5+.5(r-.5)$. Their difference, the gain in Eq.~\ref{eq:routing-option}, is $(\rho-.5)(r-.5)$. It vanishes if either read channel is at chance.

Even a wrong routing decision can have a small effect on accuracy near a value boundary. The loss from choosing $S$ is $\max(0,-D)$, where $D$ is the selector advantage in the moving-boundary experiment. At shared reliability .74, GPT-4.1 missed half the optima by always choosing $S$, yet the exact two-step scores removed only .55 percentage points of mean action loss across the 96 decisions. Hits reveal whether the policy crossed the right boundary. Value loss measures the practical consequence of crossing on the wrong side. 

\paragraph{Derive the moving-boundary example.}
With independent priors $(p_S,p_T,p_F)=(.5,.3,.3)$, branch reliability $.86$ and selector reliability $\rho\in[.5,1]$, a branch return is one with probability $.356$. Its posterior true-one probabilities are $129/178$ after return one and $3/46$ after return zero. Reading the selector leaves the answer prior at $.3$; reading its indicated branch next gives
\begin{equation}
 Q_2(H_0,S)=.5576+.3024\rho.
 \label{eq:selector-boundary-value}
\end{equation}
For a branch-first policy, a zero return leaves both branch posteriors below $.5$, so neither remaining read changes the optimal answer. After a one return, reading the other branch has continuation value $1441/2225$, whereas reading the selector has value $\max\{228/445,(128+189\rho)/445\}$. Weighting these conditional cases yields the exact piecewise value
\begin{equation}
 \begin{aligned}
 Q_2(H_0,T)&=Q_2(H_0,F)\\
 &=\begin{cases}
 .75696,& .5\leq\rho\leq89/105,\\
 .6288+.1512\rho,&89/105<\rho\leq1.
 \end{cases}
 \end{aligned}
 \label{eq:branch-boundary-value}
\end{equation}
The crossover lies in the first segment: $(.75696-.5576)/.3024=89/135$. A distractor-first policy is worth $.7088$ and cannot win here. These expressions follow by conditioning on read returns; the full latent-world recursion independently agrees at the crossover and on both segments. All fractions, terminating decimals and the crossover above are exact.

\paragraph{Plan with estimated values and costs.}
Let $\widehat Q(a)$ estimate $Q_b(H,a)$ at a fixed public history, with $\sup_a|\widehat Q(a)-Q_b(H,a)|\leq\epsilon$. Let $a^*\in\arg\max_a Q_b(H,a)$ and $\hat a\in\arg\max_a\widehat Q(a)$. Then
\begin{equation}
\begin{aligned}
 Q_b(H,a^*)-Q_b(H,\hat a)
 &\leq \widehat Q(a^*)+\epsilon-\widehat Q(\hat a)+\epsilon\\
 &\leq2\epsilon.
\end{aligned}
\label{eq:estimated-value-bound}
\end{equation}
A unique best action separated by more than $2\epsilon$ therefore retains its rank. An estimated winner with an estimated gap larger than $2\epsilon$ is also certified; a smaller estimated gap alone cannot certify the winner. This is a conditional value-estimation bound, not an empirical calibration guarantee. A valid uniform error bound is required at each visited state. If the policy maximizes these estimated remaining-budget values at every step and uses Bayes terminal decoding, summing the stepwise bounds in the return identity gives $V_B(H_0)-J_\mu\leq2B\epsilon$. This implication requires those bounds to hold along the policy-induced histories; it is not established by a small error on an unrelated evaluation set.

The experiments prescribe a reveal budget. If an agent can instead stop, let $c(a)$ be the cost of a read in accuracy-equivalent units. With $W_0(H)=V_0(H)$, the net value becomes
\begin{equation}
 \begin{aligned}
 Q_b^{c}(H,a)&=\mathbb{E}[W_{b-1}(H')\mid H,a]-c(a),\\
 W_b(H)&=\max\{V_0(H),\max_a Q_b^{c}(H,a)\},
 \end{aligned}
 \label{eq:cost-stop}
\end{equation}
where $H'$ is the updated history after the read.
For illustration, charge .10 for the selector and .30 for either branch in the fair switch task. With one reveal, the best net read value is $.75-.30=.45$, below the .50 value of answering now. With two reveals, reading the selector and then the active branch gives $1-.10-.30=.60$. A myopic stopping rule would forgo this sequence. The calculation uses hypothetical costs. The dense-page experiment measures a related resource trade-off. On the earlier synthetic dense-page panel, selected crops used fewer input tokens than native reading while requiring twice as many calls. This differs from the FUNSD deployment: its one-crop systems use more input tokens and two calls versus one native call (Table~\ref{tab:deployment-comparison}).